\documentclass[]{article}

\usepackage{authblk}
\usepackage[T1]{fontenc}
\usepackage[utf8]{inputenc}
\usepackage{amsmath}
\usepackage{amssymb}
\usepackage{amsfonts}
\usepackage{graphicx}
\usepackage{geometry}
\usepackage{multirow}
\usepackage{algorithm}
\usepackage{algpseudocode}
\usepackage{array}
\usepackage{fixltx2e}
\usepackage{url}
\usepackage{color}
\usepackage{xcolor}
\usepackage{scrextend}
\usepackage{tikz}
\newcolumntype{?}{!{\vrule width 1.3pt}}

\newcommand{\mbf}[1]{\mathbf{#1}}
\newcommand{\mbs}[1]{\boldsymbol{#1}}

\title{Sample size estimation for power and accuracy in the experimental comparison of algorithms.}
\author[1,2]{Felipe Campelo}
\author[1,3]{Fernanda Takahashi}
\affil[1]{Operations Research and Complex Systems Laboratory\\
  Universidade Federal de Minas Gerais\\
  Belo Horizonte 31270-010, Brazil.}
\affil[2]{Department of Electrical Engineering\\
  Universidade Federal de Minas Gerais, Brazil.
  E-mail: fcampelo@ufmg.br}
\affil[3]{Graduate Program in Electrical Engineering\\
  Universidade Federal de Minas Gerais, Brazil.
  E-mail: fernandact@ufmg.br}
  
\begin{document}
\begin{titlepage}
	\vspace*{3cm}
	\begin{center}
		\huge {Sample size estimation for power and accuracy in the experimental comparison of algorithms} 
	\end{center}
	\vspace{2cm} 
	\begin{center} 
		\Large {Felipe Campelo, Fernanda Takahashi} \\[3pt]  
		\textit{Universidade Federal de Minas Gerais\\
			Belo Horizonte 31270-010, Brazil.} \\ 
		\today
		
		\vspace{2cm}
		\large{\textbf{Disclaimer}\\This is a pre-publication author version, which does not incorporate the final reviews performed on the manuscript.\\
			While this version has no known technical mistakes, readers should (whenever possible) use the final version, which incorporates supplemental materials and was published in the \textbf{Journal of Heuristics}. The final version is available at:
			
			\vspace{.5cm}
			\url{https://link.springer.com/article/10.1007%2Fs10732-018-9396-7}}
	\end{center}
\end{titlepage}

\clearpage

\maketitle

\begin{abstract}
Experimental comparisons of performance represent an important aspect of research on optimization algorithms. In this work we present a methodology for defining the required sample sizes for designing experiments with desired statistical properties for the comparison of two methods on a given problem class. The proposed approach allows the experimenter to define desired levels of accuracy for estimates of mean performance differences on individual problem instances, as well as the desired statistical power for comparing mean performances over a problem class of interest. The method calculates the required number of problem instances, and runs the algorithms on each test instance so that the accuracy of the estimated differences in performance is controlled at the predefined level. Two examples illustrate the application of the proposed method, and its ability to achieve the desired statistical properties with a methodologically sound definition of the relevant sample sizes.
\end{abstract}

\section{Introduction}
Research on optimization algorithms, particularly heuristics, tends to rely heavily on experimental results to evaluate and compare different methods, as well as to measure the impact of algorithmic modifications. This central role of experimentation has generated an ongoing interest in the definition of adequate experimental protocols and the use of inferential procedures for comparing two or more algorithms, either on single or multiple problem instances \cite{Barr1995,McGeoch1996,Hooker1996,CoffinSaltzman2000,Johnson2002,Yuan2004,Demsar2006,YuanGallagher2009,Birattari2004,Birattari2007,Bartz-Beielstein2006,Bartz-Beielstein2010,Garcia2008,Garcia2010,Derrac2011,Derrac2014,Carrano2011,Krohling2015,COCO2016,Benavoli2014}. Despite these improvements, however, most works dealing with the proposal and application of statistical protocols for algorithm comparison still lack a solid discussion on the required sample sizes -- number of instances, number of repeated runs -- to obtain an experiment with predefined statistical properties. Other related themes are also rarely found in the literature: discussions on the statistical power and accuracy of parameter estimation; on the observed effect sizes and their practical significance; and on the problem class(es) for which the conclusions of a particular experiment can be extended.

Regarding sample sizes, the standard approach in most cases has been one of maximizing the number of instances, limited only by the computational budget available \cite{Barr1995,Garcia2008,Garcia2009,Derrac2011,Amo2012}; and of using ``standard'' values for the number of repeated runs (usually 30 or 50, with occasional recommendations of using 100 or more). While the probability of detecting a difference in performance between algorithms does indeed increase as the effective sample size grows \cite{CoffinSaltzman2000,Bartz-Beielstein2006}, this does not imply that sample sizes need to be arbitrarily large for an experiment to yield high-quality conclusions, or that a large sample size can substitute a well-designed experiment \cite{Lenth2001,Chow2003,Bausell2006,Mathews2010}. 

A first point against the approach of blindly maximizing sample size is that statistical analyses made with moderately-sized samples, or even small ones, can be as useful and convincing as ones made with very large samples \cite{CoffinSaltzman2000,Mathews2010}, provided that the experimental design is adequate, the test assumptions are valid, and the sample is representative -- i.e., the test instances used represent a typical (ideally random) sample of the problem class for which conclusions are to be drawn. While there are works focused on small sample sizes \cite{YuanGallagher2009}, the use of large test sets is constantly reinforced despite of computational difficulties in some cases, e.g., applied optimization based on numeric models \cite{Mori2015}. 

Second, and more critically, arbitrarily large sample sizes allow statistical procedures to detect even minuscule differences, which may lead to the wrongful interpretation that effects of no practical consequence are strongly significant \cite{Mathews2010,BartzBeielstein2005}. Using arbitrarily large samples increases the probability that such small differences (in many cases due to implementation details or subtle differences in tuning) influence the conclusions and recommendations of a given study, suggesting practical superiority where none exists. Of course this problem can be alleviated by defining a \textit{minimally relevant effect size} (MRES) (also known as the \textit{smallest difference of practical importance}) prior to the experiment, but in practice this is rarely, if ever, found in the literature on experimental comparison of algorithms.

Assuming that we are only interested in detecting effects that may have some practical consequence -- i.e., differences larger than a given MRES -- it is possible to define the smallest number of observations required to obtain predefined statistical power and accuracy, instead of arbitrarily increasing the sample size. In the case of algorithmic comparisons, two types of sample sizes are of particular interest: (i) the number of \textit{repetitions}, i.e., repeated runs of an algorithm on a given problem instance under different initial conditions, such as seed values for (pseudo-)random number generators, or the initial candidate solution for the search; (ii) the number of \textit{replicates}, i.e., the amount of test instances used in the experiment. 

There is no established rule on how to estimate either value in the literature on experimental comparison of algorithms. Jain \cite{Jain1991} discusses a simple, standard formula for determining the number of repetitions for a desired confidence interval half-width, given the value of the sample standard deviation for a single algorithm on a single instance. Coffin and Saltzman \cite{CoffinSaltzman2000} mention the idea of sample size calculations, but provide no further discussion on the topic. Czarn \textit{et al.} \cite{Czarn2004} discusses power and sample size in the context of the number of repeated runs, when the experiment is intended at uncovering differences of algorithms on a single problem instance. Their work also suggest a sequential inference procedure, iteratively increasing the sample size and re-testing until an effect is found or a statistical power of $0.8$ is reached for a given MRES, but fails to adequately correct for the increase in type-I errors due to multiple hypothesis testing \cite{Shaffer1995,Botella2006}. Ridge \cite{Ridge2007} and Ridge and Kudenko \cite[Ch. 11]{Bartz-Beielstein2010} follow a similar approach, and also do not correct the significance level for multiple hypothesis testing.

Birattari \cite{Birattari2004,Birattari2009} correctly advocates for a greater focus on the number of instances than on repetitions, showing that the optimal allocation of computational effort, in terms of accuracy in the estimation of mean performance for a given problem class, is to maximize the amount of test instances, running each algorithm a single time on each instance if needed. This approach, however, yields very little information on the specific behavior of each  algorithm on each instance, precluding further analysis and investigation of specific aspects of algorithmic performance. It also does not take into account the questions of desired statistical power, sample size calculation, or the definition of a MRES.

Bartz-Beielstein \cite{BartzBeielstein2005,Bartz-Beielstein2006} discusses the perils of using a sample size that is too large, in terms of the increase in spurious ``significant'' results, and also provides some comments on the need to detect effect sizes of practical importance. Finally, Chiarandini and Goegebeur \cite[Ch. 10]{Bartz-Beielstein2010} provide a discussion on statistical power and sample size in the context of nested linear statistical models and provide some guidelines on the choice of the number of instances and number of repeated runs based on the graphical analyses of power curves.

In this work we present an algorithmic methodology for defining both the amount of test instances and the number of repeated runs required for an experiment to obtain desired statistical properties. More specifically, the proposed experimental framework aims at guaranteeing (i) a predefined accuracy in the estimation of paired differences of performance between two algorithms on any given instance, based on optimal sample size ratios; and (ii) a given statistical power to detect differences larger than a predefined MRES, when comparing the performance of two algorithms on a given problem class. 

An important aspect of the methodology proposed in this paper is the fact that the estimation of each relevant sample size (number of instances, number of within-instance repeated runs) can be done independently of the other: as an example, researchers testing their methods on a predefined set of benchmark instances (as is common practice in comparative performance assessment) can employ the full problem set, and use the methodology only to define the required number of runs for each algorithm on each instance, as well as the statistical power achievable by their experiment. Conversely, the method also works if the researcher wants to maintain the number of repetitions fixed (e.g., at 30 runs/algorithm/instance), and only estimate the smallest number of instances required for a given test.  Moreover, the estimation of the minimal sample sizes required for designing experiments with predefined statistical properties does not preclude the use of other methods of analysis, such as graphical profiling or the investigation of performance on individual instances. On the contrary, the methodology proposed in this paper can be easily extended to other approaches, providing researchers with a richer and more methodologically sound approach for designing experiments for performance evaluation and comparison of algorithms.

The remainder of this paper is organized as follows: in the next section we formally define the \textit{algorithm comparison problem} that is tackled in this work, namely that of comparing the (mean) performance of two algorithms on a given problem class, based on information gathered from a representative subset of instances. The definitions provided in Section \ref{sec:algcomp} can be later extended to instantiate a number of other comparisons, and will hopefully provide a basis onto which more sophisticated comparison methodologies will be eventually built.

Section \ref{sec:stats} provides the main statistical concepts and definitions needed to introduce the proposed method for calculating the required sample sizes for experiments involving algorithms. The proposed method is introduced in Section \ref{sec:caiser}, where sample size calculations are detailed for two specific cases, namely that of simple differences and percent differences in the mean performance of two algorithms. The application of the proposed methodology is illustrated in two examples of application discussed in Section \ref{sec:examples}. Finally, concluding remarks and possibilities of continuity are discussed in Section \ref{sec:conclusions}.

\section{The Algorithm Comparison Problem}\label{sec:algcomp}

Let $\Gamma = \left\{\gamma_1, \gamma_2,\dotsc\right\}$ represent a \textit{problem class} consisting of a set of (possibly infinitely many) problem instances $\gamma_j$ which are of interest as a group (e.g., the set of all possible TSP instances within a given size range); and let $\mathcal{A} = \left\{a_1, a_2, \dotsc\right\}$ denote a set of algorithms capable of returning tentative solutions to each instance $\gamma_j\in\Gamma$.\footnote{Throughout this work we refer to an algorithm as the full structure with specific parameter values, i.e., to a completely defined instantiation of a given algorithmic framework.} In this work, we are interested in comparing the performance of two algorithms $a_1, a_2\in\mathcal{A}$ as solvers for a given problem class $\Gamma$.\footnote{\ $\Gamma$ can either be explicitly known or implicitly defined as a hypothetical set for which some available test instances can be considered a representative sample.} We assume that both algorithms of interest can be run on the same subset of instances, and that any run of the algorithm returns some tentative solution, which can be used to assess the quality of that result. 

Let $\phi_j = f\left(a_1,a_2,\gamma_j\right):\mathcal{A}^2\times\Gamma\mapsto\mathbb{R}$ denote the difference in performance between algorithms $a_1,a_2$ on instance $\gamma_j$, measured according to some indicator of choice; 
and let $\Phi = \left\{\phi_j:a_1,a_2\in\mathcal{A}, \gamma_j\in\Gamma\right\}$ denote the set of these \textit{paired differences in performance} between $a_1$ and $a_2$ for all instances $\gamma_j\in\Gamma$, with $P\left(\Phi\right)$ denoting the probability density function describing the distribution of values $\phi_j\in\Phi$.

Given these definitions, the \textit{algorithm comparison problem} discussed in this work can be generally defined, given two algorithms $a_1,a_2\in\mathcal{A}$ and a problem class $\Gamma$, as the problem of performing inference about a given parameter $\theta$ of the underlying distribution $P\left(\Phi\right)$, based on information obtained by running $a_1$ and $a_2$ a certain number of times on a finite sample of instances $\Gamma_S\subset\Gamma$. The parameter of interest, $\theta$, should represent a relevant quantity on which algorithms are to be compared. Common examples of parameters of interest are the mean of $P\left(\Phi\right)$, in which case the comparison problem presented here would result in the test of hypotheses on the paired difference of means (performed using, e.g., a \textit{paired t-test}); or the median, in which case we could use the \textit{Wilcoxon signed-rank test} or the \textit{binomial sign test} \cite{Montgomery2013}.

Finally, assume that the result of a given run of algorithm $a_i$ on instance $\gamma_j$, denoted $x_{ij}$, is subject to random variations -- e.g., due to $a_i$ being a randomized algorithm, or to randomly defined initial states in a deterministic method -- such that $x_{ij}\sim X_{ij}$, where $X_{ij}$ is the underlying random variable associated with the distribution of performance values for the pair $\left(a_i,\gamma_j\right)$. 

Notice that these assumptions, which represent the usual case for the majority of experimental comparisons of algorithms, mean that there are two sources of uncertainty that must be considered when trying to address the algorithm comparison problem. First, there is the uncertainty arising from the fact that we are trying to answer questions about a population parameter $\theta$ based on a limited sample, which is the classical problem of statistical inference. 
The second source of variability is the uncertainty associated with the estimation of $\phi_j$ from a finite number of runs.

These two components of the total variability of the results to be used for comparing two algorithms influence the statistical power of any inferential task to be performed on the value of  $\theta$. To control these influences there are two types of sample sizes that need to be considered:
\begin{itemize}
\item The number of repeated runs (\textit{repetitions}), i.e., how many times each algorithm $a_i$ needs to be run on each instance $\gamma_j$. These sample sizes, which will be denoted $n_{ij}$, can be used to control the accuracy of estimation of $\phi_j$ and, to a lesser extent, contribute to the statistical power of the comparison;
\item The number of problem instances used in the experiment (\textit{replicates}), also called here the \textit{effective sample size}. This value, which will be denoted $N = \left|\Gamma_S\right|$, can be used to more directly set the statistical power of the comparison at a desired level.
\end{itemize}

In this work we focus on comparisons of mean performance, with simple extensions to the testing of medians. The specifics of this particular case are discussed next.

\subsection{Comparison of mean performance}
\label{sec:algcomp:H0H1}
When comparing two algorithms in terms of their mean performance over a given problem class of interest, we are generally interested in performing inference on the value of $\theta = \mu_D = E\left[P\left(\Phi\right)\right]$. In this case, the statistical hypotheses to be tested, if we are interested in  simply  investigating the existence of differences in mean performance between the two algorithms, regardless of their direction, are: 
%
\begin{equation}
\label{eq:HoH1a}
\begin{matrix}
H_0: &\mu_D = \mu_0\\
H_1: &\mu_D \neq \mu_0;
\end{matrix}
\end{equation}

\noindent or, if we are interested in specifically determining whether algorithm $a_2$ (e.g., a proposed approach) is superior to $a_1$ (e.g., a state-of-the-art approach) in terms of mean performance over the problem class of interest\footnote{The direction of the inequalities in \eqref{eq:HoH1b} will depend on the type of performance measure used, i.e., on whether \textit{larger = better} or vice versa.}, 
\begin{equation}
\label{eq:HoH1b}
\begin{matrix}
H_0: &\mu_D \geq \mu_0\\
H_1: &\mu_D < \mu_0.
\end{matrix}
\end{equation}


The value of $\mu_0$ in \eqref{eq:HoH1a} and \eqref{eq:HoH1b}, i.e., the mean of the paired differences of performance under the null hypothesis $H_0$, is commonly set as $\mu_0 = 0$ when comparing algorithms, reflecting the absence of prior knowledge of differences in performance between the two algorithms compared.

As mentioned earlier in this section, there are two types of sample sizes that need to be considered for comparing algorithms: the number of within-sample repetitions for each algorithm, and the number of instances to be employed. In Section \ref{sec:caiser} we present a methodology for calculating these two sample sizes for the algorithm comparison problem defined in this section. Prior to describing the method, however, it is important to review some relevant statistical concepts that provide the basis for the proposed approach.

\section{Relevant Statistical Concepts}
\label{sec:stats}
In this section we discuss the main statistical concepts needed to introduce the proposed methodology for sample size estimation. More specifically, we provide a brief overview on the statistical error rates associated with the test of hypotheses \cite{Montgomery2013}, and discuss the concept of \textit{statistical power} and its relationship with the \textit{effect size} \cite{Grissom2012,Mathews2010} in the comparison of algorithms. We employ these concepts to define the \textit{minimally relevant effect size} (MRES), also known as the \textit{smallest difference of practical significance}, which is essential for estimating the smallest number of instances required for the comparison of algorithms. Finally, we present a brief discussion about accuracy of parameter estimation, which will be important as a basis for calculating the number of runs required for each algorithm on each instance.

\subsection{Statistical errors and effect size}
Statistical inference can be seen as a methodology for deciding between two competing statements regarding a given populational parameter, based on an incomplete sample, with a quantifiable degree of confidence. This process is subject to two types of statistical errors \cite{Montgomery2013}: Type I, which quantifies the probability of incurring in a \textit{false positive}, i.e., wrongly rejecting a true null hypothesis; and Type II, which represents the probability of a \textit{false negative} -- failing to reject a null hypothesis that is false. The Type I error rate of an inferential procedure is quantified by the \textit{significance level} $\alpha$ (or the \textit{confidence level}, $1-\alpha$), which depends only on the null hypothesis and can be directly controlled by the experimenter. On the other hand, the Type II error rate $\beta$ (or, equivalently, the statistical power $\pi = 1 - \beta$), depends on several parameters, some of which cannot be so easily controlled. 

Given a pair of hypotheses of the form \eqref{eq:HoH1a} or \eqref{eq:HoH1b}, the power of a statistical test is related to two controllable quantities, namely the significance level $\alpha$ and the sample size $N$; and one uncontrollable one, namely the ratio 
\begin{equation}
\label{eq:CohenD}
d = \frac{\left|\mu - \mu_0\right|}{\sigma} = \frac{\left|\delta\right|}{\sigma},
\end{equation}

\noindent where $\sigma$ denotes the standard deviation of $P\left(\Phi\right)$, and $\delta$ the difference between the true mean of $P\left(\Phi\right)$ and the value stated in the null hypothesis, $\mu_0$. The difference $\delta$ is called the \textit{simple effect size}, and  $d$ is known as the \textit{standardized effect size} or \textit{Cohen's d} coefficient \cite{Grissom2012}. 

Generally speaking, statistical power can be interpreted as the \textit{sensitivity} of a test to a given effect size, i.e., its probability of detecting deviations from the null hypothesis at or above a certain magnitude. All other quantities being equal, the power of a test increases as $\alpha, N$ and $d$ increase. The significance level is usually set \textit{a priori} by the experimenter, and the sample size is also commonly controllable (at least up to a certain value), but the true value of the effect size is usually unknown (otherwise there would be no need for inference). However, the fact that power increases with $d$ means that, if we define a \textit{minimally relevant effect size} $d^* = \left|\delta^*\right|/\sigma$ -- i.e., the smallest value of $d$ that the experimenter is interested in detecting -- and design the experiment to have a desired power for the limit case where $d = d^*$, then the test will have greater power for any effect size $d>d^*$. Values smaller than $d^*$ will of course result in lower power, but by definition these values are deemed uninteresting, and as such failure to detect them is of no practical consequence. Setting an adequate $d^*$ means that the experimenter can calculate the necessary sample size $N$ to achieve the desired power level. In the next section we show how this can be done for statistical tests on the value of $\mu_D$.

Finally, while the definition of $d^*$ (or of $\delta^*$, if a reasonable upper estimate of $\sigma$ can be provided) is important for determining the required sample size for a given experiment, we argue that its importance goes beyond this aspect. By forcing the experimenter to define which differences would have practical relevance \textit{prior to the experiment}, the MRES helps to avoid well-known problems associated with the exclusive use of the p-value in statistical inference \cite{Nuzzo2014,BartzBeielstein2005}. In short, it provides a \textit{practical relevance} dimension to the usual tests of statistical significance.

\subsection{Parameter estimation and accuracy}
\label{sec:accuracy}
One of the most common uses of statistics is parameter estimation, i.e., the use of information contained in a finite sample to estimate, with a certain accuracy, the value of a given parameter. For any parameter $\theta$, the usual notation for its point estimator is $\widehat{\Theta}$, and a specific value of this point estimator is a point estimate, $\widehat{\theta}$ \cite{Montgomery2013}. Two common examples of point estimators, which have their own specific notations, are the sample mean and the sample standard deviation,
\begin{align}
\bar{X} &= \widehat{\mu} = \frac{1}{n}\sum\limits_{i=1}^{n}X_i, \label{eq:sampleXbar}\\
S &= \widehat{\sigma} = \sqrt{\frac{1}{n-1}\sum\limits_{i=1}^{n}\left(X_i - \bar{X}\right)^2}. \label{eq:sampleS}
\end{align}


While point estimators return the value of greatest likelihood for a parameter given a sample, their values are also subject to uncertainties due to the randomness of their inputs. More specifically, a point estimator $\widehat{\Theta}$ has a given \textit{sampling distribution} \cite{Montgomery2013}, which is a function of populational parameters and the sample size used in its calculation. The sample mean, for instance, has a distribution $P\left(\bar{X}\right)$ with mean $E\left[\bar{X}\right] = E\left[X\right] = \mu$ and variance $V\left[\bar{X}\right] = V\left[X\right]/n = \sigma^2/n$. 

Given these aspects of parameter estimation, an important point to consider is the \textit{accuracy} of parameter estimates. A simple way of measuring this accuracy is using the \textit{standard error} $se_{\widehat{\theta}}$, which represents the standard deviation of the sampling distribution of the estimator \cite{Montgomery2013}. For the sample mean, for instance, the standard error is given as $se_{\bar{X}} = \sigma/\sqrt{n}$, and can be generally interpreted as analogous to a ``measurement error'' of the parameter being estimated, in this case the true mean $\mu$. Since in most cases the  populational standard deviation is not known, it must be estimated from the data, which results in the calculation of the \textit{sample standard error},
\begin{equation}
\label{eq:sehat}
\widehat{se}_{\bar{X}} =s / \sqrt{n},
\end{equation}

Notice that it is straightforward to solve for $n$ in \eqref{eq:sehat}, which allows us to predefine a desired level of accuracy (i.e., a desired upper limit for $\widehat{se}_{\bar{X}}$) for the estimation and calculate the required sample size to obtain it. Since we need some data to estimate $s$ in the first place, an iterative approach can be used for this calculation, as will be presented in Section \ref{sec:repetitions}.

\section{Proposed Approach}
\label{sec:caiser}
Given the definitions provided in the preceding sections, we present a methodology for estimating the relevant sample sizes for the algorithm comparison problem described in Section \ref{sec:algcomp}, that is, the comparison of two algorithms in terms of their mean paired differences of performance over instances belonging to a given problem class. More specifically, we describe (i) an algorithmic approach to iteratively sample each algorithm on each problem instance (i.e., \textit{repetitions}) with sample size ratios close to theoretical optimal values, so that a predefined accuracy in the estimation of each $\phi_j$ is obtained; and (ii) specific formulas for determining the required number of instances (i.e., \textit{replicates}), so that a desired power level can be achieved for a predefined MRES. 

As mentioned in the Introduction, it is important to highlight here that the two main results of the proposed methodology, namely the estimation of the number of instances and number of within-instance replicates, are independent: researchers can employ the two calculations separately if desired or required by the specifics of a particular experiment. For instance, it is common for certain application domains to have predefined test sets composed of heterogeneous instances, aimed at testing the behavior of algorithms on a variety of possible situations. In these cases the researcher may wish to employ the full set of available test instances (assuming computational time is not an issue), but he or she can still employ the proposed methodology to: (i) determine the number of runs for each algorithm on each instance (see Section \ref{sec:repetitions} below), and (ii) determine the statistical properties of the experiment in terms of the power to detect differences of a given magnitude (see Section \ref{sec:fixedSS}). In any case, the application of the principles discussed in this work can aid the research to design and perform comparative experiments with increased statistical soundness.

Finally, for readers who want to skip the derivations of the method, a short summary is available in the Supplementary Materials, together with a quick use guide.

\subsection{Estimating the number of repetitions}
\label{sec:repetitions}
The proposed strategy to calculate the number of runs of each algorithm $a_i$ on a given instance $\gamma_j$, (i.e., the number of repetitions, $n_{ij}$)  consists in iteratively increasing the number of observations of each algorithm until the standard error of $\widehat{\phi}_j$ (the estimate of the difference in performance between the two algorithms for instance $\gamma_j$) falls below a given threshold. While the specifics of standard error estimation depend on which statistic is being used to quantify the difference in performance, the accuracy of estimation improves as the sample sizes $n_{1j}$ and $n_{2j}$ are increased. This allows us to define the problem of estimating the number of runs of algorithms $a_1,a_2$ on a given instance $\gamma_j$ as that of \textit{finding the smallest total sample size, $n_{1j} + n_{2j}$, such that the standard error of $\widehat{\phi}_j$ falls below a desired accuracy threshold $se^*$.} 

Notice that unlike the usual practice in the experimental comparison of algorithms, the solution for this problem will almost always result in different numbers of runs of $a_1$ and  $a_2$ on any given instance. This is a consequence of the fact that distinct algorithms will present different variances of performance within any instance, which means that their contributions to the standard error of any estimator used to quantify the paired differences in performance will be unequal. In general, the larger-variance algorithm will need a larger sample size, as will be made clear in this section. Notice, however, that the method presented in this section will work perfectly well if the experimenter forces equal sample sizes (which can be done by a small, trivial modification of Algorithm \ref{alg:reps_abs}), although the total number of runs in this case may be larger than necessary.

In what follows we provide the derivation of the optimal sample sizes for two specific cases of $\phi_j$, namely the \textit{simple} and the \textit{percent} difference between two means. The derivations are performed assuming that the conditions for the Central Limit Theorem (CLT) are met \cite{Montgomery2013}, which means that the sampling distributions of the means are approximately normal. An alternative approach, which does not need to comply with this particular set of assumptions (at the cost of increased computational costs) involves the use of resampling strategies such as the Bootstrap \cite{Efron1994:book}, which is discussed in Section \ref{sec:nonpar}.

\subsubsection{Using the Simple Difference of Two Means}
\label{sec:abs_diff}
Assume that we are interested in using the simple difference of mean performance between algorithms \(a_1,a_2\) on each instance as our
values of \(\phi_j\). 
In this case we define \(\phi_j = \mu_{2j} - \mu_{1j}\), for which the sample estimator is given by
\begin{equation}
\label{eq:diff_ab}
\widehat{\phi}_j^{(1)} = \widehat{\Delta\mu} = \widehat{\mu}_{2j} - \widehat{\mu}_{1j} = \bar{X}_{2j} - \bar{X}_{1j},
\end{equation}

\noindent where \(\bar{X}_{ij}\) is the sample mean of algorithm \(a_i\) on instance \(\gamma_j\). Let the distribution of performance values of algorithm \(a_i\) on instance \(\gamma_j\) be expressed as an (unknown) probability density function with expected value \(\mu_{ij}\) and variance \(\sigma^2_{ij}\), i.e., \[x_{ij} \sim X_{ij} = \mathcal{P}\left(\mu_{ij},\sigma^2_{ij}\right).\]

Assuming that the conditions of the Central Limit Theorem hold, we expect \(\bar{X}_{ij} \sim \mathcal{N}\left(\mu_{ij},\sigma^2_{ij}/n_{ij}\right)\) and, consequently,
\begin{equation}
\label{eq:D_abs}
\widehat{\phi}_j^{(1)}\sim \mathcal{N}\left(\mu_{2j} - \mu_{1j},\frac{\sigma^2_{1j}}{n_{1j}} + \frac{\sigma^2_{2j}}{n_{2j}}\right).
\end{equation}

By definition, the standard error of \(\widehat{\phi}_j^{(1)}\) is the standard deviation of this sampling distribution of the estimator, \[se_{\widehat{\phi}_j^{(1)}} = \sqrt{\sigma^2_{1j}n_{1j}^{-1} + \sigma^2_{2j}n_{2j}^{-1}}.\]

Given a desired upper limit for the standard error, \(se^*\), the optimal sample sizes for the two algorithms \(a_1,a_2\) on instance \(\gamma_j\) can be obtained by solving the optimization problem defined as
\begin{equation}
\label{eq:opt_repsize}
\begin{split}
\mbox{Minimize: }& f(\mathbf{n}_j)  = n_{1j} + n_{2j},\\
\mbox{Subject to: }& g(\mathbf{n}_j) = se_{\widehat{\phi}_j^{(1)}} - se^* \leq 0.
\end{split}
\end{equation}

This problem can be solved analytically using the Karush-Kuhn-Tucker
(KKT) optimality conditions,
\begin{equation}
\label{eq:opt_sys}
\begin{aligned}
\nabla f(\mathbf{n}_j)  + \beta\nabla g(\mathbf{n}_j) &= 0,\\
\beta g(\mathbf{n}_j) &= 0,\\
\beta&\geq 0.
\end{aligned}
\end{equation}

%
%
%

The solution of \eqref{eq:opt_sys} for the objective and constraint functions in \eqref{eq:opt_repsize} yields the optimal ratio of sample sizes,
\begin{equation}
\label{eq:opt_ratio}
r_{opt} = \frac{n_{1j}}{n_{2j}} = \frac{\sigma_{1j}}{\sigma_{2j}},
\end{equation}

\noindent which means that algorithms \(a_1\) and \(a_2\) must be
sampled on instance \(\gamma_j\) in direct proportion to the standard
deviations of their performances on that instance. The result in
\eqref{eq:opt_ratio} is known in the statistical literature \cite{Mathews2010} as the \emph{optimal allocation of resources} for the estimation
of confidence intervals on the simple difference of two means. 
%

Since the populational variances $\sigma_{1j}^2,\sigma_{2j}^2$ are usually unknown, their values need to be estimated from the data. This results in the sample standard error,
\begin{equation}
\label{eq:se_abs}
\widehat{se}_{\widehat{\phi}_j^{(1)}} = \sqrt{s^2_{1j}n_{1j}^{-1} + s^2_{2j}n_{2j}^{-1}}.
\end{equation}

A good approximation of the optimal ratio of sample sizes can be similarly obtained by replacing $\sigma_{ij}$ by $s_{ij}$ in
\eqref{eq:opt_ratio}. This requires that an initial sample size \(n_0\) be obtained for each algorithm, to calculate initial estimates of \(s_{1j},s_{2j}\)\footnote{The definition of an
initial value of \(n_0\) also helps increasing the probability that the
sampling distributions of the means will be
approximately normal.}, suggesting the iterative procedure described in Algorithm \ref{alg:reps_abs}, where \verb|Sample|$\left(a_i,\gamma_j,n~\mbox{times}\right)$ means to obtain $n$ observations of algorithm $a_i$ on instance $\gamma_j$.  To prevent an explosion of the number of repetitions in the case of poorly specified threshold values $se^*$ or particularly high-variance algorithms, a maximum budget $n_{max}$ can also be defined for the sampling on a given problem instance.
\begin{algorithm}
	\caption{Sample algorithms on one instance.}
	\label{alg:reps_abs}
	\begin{algorithmic}[1]
		\Require Instance $\gamma_j$; Algorithms $a_1,a_2$; accuracy threshold $se^*$; initial sample size $n_0$; maximum sample size $n_{max}$.
		\State $\mbf{x}_{1j}\leftarrow$ Sample$\left(a_1,\gamma_j,n_0~\mbox{times}\right)$
		\State $\mbf{x}_{2j}\leftarrow$ Sample$\left(a_2,\gamma_j,n_0~\mbox{times}\right)$
		\State $n_{1j}\leftarrow n_0$\;
		\State $n_{2j}\leftarrow n_0$\;
		\State Calculate $\widehat{se}$ using \eqref{eq:se_abs} or \eqref{eq:FiellerSE_nocov} or Algorithm \ref{alg:bootSE}\;
		\While{$\left(\widehat{se} > se^*\right)$ \& $\left(n_{1j}+n_{2j}< n_{max}\right)$}
		\State Calculate $r_{opt}$ using \eqref{eq:opt_ratio} or \eqref{eq:opt_ratio2}\;
		\If{$\left(n_{1j}/n_{2j} < r_{opt}\right)$}
		\State $x\leftarrow$ Sample$\left(a_1,\gamma_j,1~\mbox{time}\right)$\;
		\State $\mbf{x}_{1j}\leftarrow [\mbf{x}_{1j}, x]$\;
		\State $n_{1j}\leftarrow n_{1j} + 1$\;
		\Else
		\State $x\leftarrow$ Sample$\left(a_2,\gamma_j,1~\mbox{time}\right)$\;
		\State $\mbf{x}_{2j}\leftarrow [\mbf{x}_{2j}, x]$\;
		\State $n_{2j}\leftarrow n_{2j} + 1$\;
		\EndIf
		\State Calculate $\widehat{se}$ using \eqref{eq:se_abs} or \eqref{eq:FiellerSE_nocov} or Algorithm \ref{alg:bootSE}\;
		\EndWhile
		\vspace{.10cm}
		\State\Return $\mbf{x}_{1j},~\mbf{x}_{2j}$\;
	\end{algorithmic}
\end{algorithm}

After performing the procedure shown in Algorithm \ref{alg:reps_abs}, the estimate $\widehat{\phi}_j$ can be calculated using the vectors of observation $\mbf{x}_{1j}$ and $\mbf{x}_{2j}$ into \eqref{eq:diff_ab} or \eqref{eq:diff_perc}, depending on the type of difference used.

\subsubsection{Using the Percent Difference of Two Means}
\label{sec:perc_diff}
While the approach of defining \(\phi_j\) as the simple difference
between the means of algorithms \(a_1,a_2\) on a given instance
\(\gamma_j\) is certainly useful, it may be subject to some difficulties. In
particular, defining a single precision threshold \(se^*\) for problem classes
containing instances with vastly different scales can be problematic and
lead to wasteful sampling. In these cases, it is generally more
practical and more intuitive to define the differences in performance
\(\phi_j\) as the \textit{percent mean gains} of algorithm \(a_2\) over
\(a_1\). In
this case we define\footnote{Considering a comparison where larger is better.}
\(\phi_j = \left(\mu_{2j} - \mu_{1j}\right)/\mu_{1j}\), for which the
sample estimator is 
\begin{equation}
\label{eq:diff_perc}
\widehat{\phi}_j^{(2)} = \widehat{\Delta\mu}_{(\%)} = \frac{\bar{X}_{2j} - \bar{X}_{1j}}{\bar{X}_{1j}} = \frac{\widehat{\phi}_j^{(1)}}{\bar{X}_{1j}}
\end{equation}

For this definition to be used we need to consider an
additional assumption, namely that
\(P(\bar{X}_{1j}\leq 0) \rightarrow 0\) (which is guaranteed, for
instance, when objective function values are always strictly positive,
which is common in several problem
classes).\footnote{If this assumption cannot be guaranteed, the use of percent differences is not advisable, and the researcher should instead perform comparisons using the simple differences.}
The distribution of \(\phi_j^{(1)}\) is given in \eqref{eq:D_abs}, which
means that under our working assumptions \(\widehat{\phi}_j^{(2)}\) is
distributed as the ratio of two independent normal
variables.\footnote{The independence between $\phi_j^{(1)}$ and $\bar{X}_{1j}$ is guaranteed as long as $X_{1j}$ and $X_{2j}$ are independent.}

A commonly used estimator of the standard error of
\(\widehat{\phi}_j^{(2)}\) is based on confidence interval derivations
by Fieller \cite{Fieller1954}. Considering the assumption that
\(P(\bar{X}_{1j}\leq 0) \rightarrow 0\), a simplified form of Fieller's
estimator can be used \cite{Franz2007}, which provides good coverage
properties. Under balanced sampling, i.e., \(n_{1j} = n_{2j} = n_j\),
the standard error is given as
\begin{equation}
\label{eq:FiellerSE}
\widehat{se}_{\widehat{\phi}_j^{(2)}} = \left|\widehat{\phi}_j^{(2)}\right| \left[ \frac{s^2_{1j}/n_j}{\bar{x}^2_{1j}} + \frac{\left(s^2_{1j}/n_j + s^2_{2j}/n_j\right)}{\left(\widehat{\phi}_j^{(1)}\right)^2}+ \frac{2}{n_j}\frac{cov\left(\mathbf{x}_{1j},\left(\mathbf{x}_{2j}-\mathbf{x}_{1j}\right)\right)}{\widehat{\phi}_j^{(1)}\bar{x}_{1j}} \right]^{1 / 2},
\end{equation}

\noindent where \(\mathbf{x}_{ij}\in \mathbb{R}^{n_j}\) represents the
vector of observations of algorithm \(a_i\) on instance \(\gamma_j\);
and \(cov\left(\cdot,\cdot\right)\) is the sample covariance of two
vectors.

Under the assumption of within-instance independence, i.e., that
\(X_{1j}\) and \(X_{2j}\) are independent, the expected value of
covariance will be close to zero, allowing us to disregard the covariance
term in \eqref{eq:FiellerSE}. This offers two advantages: first, it
simplifies calculations of the standard error, particularly for larger
sample sizes. Second, and more importantly, it allows us to consider
unbalanced sampling, as we did for the case of simple differences, 
which can lead to gains in efficiency. Removing
the covariance term, replacing the \(n_j\) dividing each sample standard
deviation by the corresponding \(n_{ij}\) and simplifying
\eqref{eq:FiellerSE} results in
\begin{equation}
\label{eq:FiellerSE_nocov}
\widehat{se}_{\widehat{\phi}_j^{(2)}} = \left|\widehat{\phi}_j^{(2)}\right|\sqrt{c_1n_{1j}^{-1} + c_2n_{2j}^{-1}},
\end{equation}

\noindent with
\begin{equation}
\label{eq:SE_nocov_c}
\begin{split}
c_1 &= s^2_{1j}\left[\left(\widehat{\phi}_j^{(1)}\right)^{-2} +\left(\bar{x}_{1j}\right)^{-2}\right];\\
c_2 &= s^2_{2j}\left(\widehat{\phi}_j^{(1)}\right)^{-2}.
\end{split}
\end{equation}

The problem of calculating the smallest total sample size required to
achieve a desired accuracy is equivalent to the one stated in
\eqref{eq:opt_repsize} (substituting
\(\widehat{se}_{\widehat{\phi}_j^{(1)}}\) by
\(\widehat{se}_{\widehat{\phi}_j^{(2)}}\) in the constraint function) and can be solved in a similar manner to yield the optimal ratio of sample sizes in the case of percent differences,
%
%
%
%
%
%
\begin{equation}
\label{eq:opt_ratio2}
r_{opt} = \frac{n_{1j}}{n_{2j}} = \sqrt{\frac{c_1}{c_2}} = \frac{s_{1j}}{s_{2j}}\sqrt{1+\left(\widehat{\phi}_j^{(2)}\right)^2}
\end{equation}


The expressions in \eqref{eq:FiellerSE_nocov} and \eqref{eq:opt_ratio2} can be used directly into Algorithm \ref{alg:reps_abs}, so that the adequate sample sizes for obtaining an estimate $\widehat{\phi}_j^{(2)}$ with a standard error controlled at a given threshold $se^*$ can be iteratively generated.

%
%

\subsection{Estimating the number of instances}
\label{sec:replicates}

As described in Section \ref{sec:algcomp}, the algorithm comparison problem treated in this work naturally induces a paired design \cite{Montgomery2013}, which allows instance effects to be modeled out. Here we discuss the definition of the number of instances required for the experiment to obtain the desired statistical properties, namely a power of at least $\pi^* = 1-\beta^*$ to detect differences equal to or greater than a minimally relevant effect size $d^*$ at a predefined significance level $\alpha$.

Before we proceed it is important to highlight that, since the hypotheses of interest concern the expected value of a distribution defined over the set of paired differences in performance, $\Phi$, the \textit{independent observations} to be used in the test of this hypotheses are the individual values $\phi_j$ (or, more accurately, their estimates $\widehat{\phi}_j$), and not the individual runs of the algorithms on each instance. Failure to realize this point leads to pseudoreplication \cite{Hurlbert1984,Lazic2010}, i.e., the calculation of test statistics under falsely inflated degrees-of-freedom, with a consequent loss of control over the statistical error rates of the tests.

Under the assumption that the sampling distributions of means for the paired differences are approximately normal, i.e., that
\begin{equation}
\label{eq:sdm_muD}
\frac{1}{N}\sum_{\gamma_j\in\Gamma_S}\widehat{\phi}_j = \widehat{\mu}_D \sim\mathcal{N}\left(\mu_D,\sigma^2_{\Phi}/N\right),
\end{equation} 

\noindent where $N = \left|\Gamma_S\right|$ is the number of instances used in the experiment, the uniformly most powerful unbiased test for hypotheses of the forms \eqref{eq:HoH1a}--\eqref{eq:HoH1b} is the \textit{paired t-test} \cite{Montgomery2013}. The test statistic for this procedure is calculated as:
\begin{equation}
t_0 = \frac{\widehat{\mu}_D - \mu_0}{\widehat{\sigma}_{\Phi}/\sqrt{N}} = \frac{\widehat{\delta}}{\widehat{\sigma}_{\Phi}}\sqrt{N} = \widehat{d}\sqrt{N},
\end{equation}

\noindent where $\widehat{\sigma}_{\Phi}$ is the sample estimate of the total standard deviation $\sigma_{\Phi}$, and $\widehat{d}$ is the sample estimate of Cohen's $d$ coefficient \eqref{eq:CohenD}. Under $H_0$ this test statistic is distributed according to Student's t distribution with $N-1$ degrees of freedom \cite{Montgomery2013}, leading to the criterion for rejecting the null hypothesis at the $1-\alpha$ confidence level being, for hypotheses of form \eqref{eq:HoH1a}
\begin{equation}
\label{eq:t_crit_2sided}
\left|t_0\right|\geq t_{1-\alpha/2}^{(N-1)};
\end{equation} 

\noindent or, for \eqref{eq:HoH1b}, 
\begin{equation}
\label{eq:t_crit_1sided}
t_0 \leq t_{\alpha}^{(N-1)},
\end{equation} 

\noindent where $t_{q}^{(df)}$ denotes the $q$-th quantile of Student's t distribution with $df$ degrees-of-freedom \cite{Montgomery2013}. 

Under the alternative hypothesis $H_1$, $t_0$ is distributed according to a \textit{noncentral t distribution} \cite{Mathews2010} with noncentrality parameter 
\[ncp = \left(\mu_D-\mu_0\right)\sqrt{N}/\widehat{\sigma}_{\Phi} = \delta\sqrt{N}/\widehat{\sigma}_{\Phi} = d\sqrt{N}\] 

Assuming a MRES $d^* = \left|\delta^*\right|/\sigma_{\Phi}$, the power of the test is given by the integral of the noncentral t distribution with $ncp^* = d^*\sqrt{N}$  over the values of $t_0$ for which $H_0$ is rejected. For instance, for the case \eqref{eq:HoH1a} the rejection region is given in \eqref{eq:t_crit_2sided}, and the power can be calculated as
\begin{equation}
\label{eq:t.power}
\pi^* = 1 - \beta^* = 1 - \int_{t = t_{\alpha/2}^{(N-1)}}^{t_{1-\alpha/2}^{(N-1)}} \left[t_{\left|ncp^*\right|}^{(N-1)}\right]dt.
\end{equation}

The sample size for this test can then be calculated as the smallest integer such that $\pi^*$ is equal to or larger than a desired power. This leads to the formulas for the required sample size for the case of the paired t-test \cite{Mathews2010} for the two-sided alternative hypothesis \eqref{eq:HoH1a},
\begin{equation}
\label{eq:t.samplesize1}
N^* = \min~N\left| t_{1-\alpha /2}^{(N-1)} \leq t_{\beta^*; \left|ncp^*\right|}^{(N-1)}\right.,
\end{equation}

\noindent or, for the directional alternative \eqref{eq:HoH1b}, 
\begin{equation}
\label{eq:t.samplesize2}
N^* = \min~N\left| t_{1-\alpha}^{(N-1)} \leq t_{\beta^*; \left|ncp^*\right|}^{(N-1)}\right..
\end{equation}

While there are no analytical solutions for \eqref{eq:t.samplesize1}--\eqref{eq:t.samplesize2}, the calculation of these sample sizes can be easily done iteratively \cite{Mathews2010}, and is available in most statistical packages, e.g., \texttt{R} \cite{R}. Algorithm \ref{alg:caiser} summarizes the full procedure for calculating the relevant sample sizes and running the experimental comparison of mean performance of two algorithms for a given problem class. As mentioned earlier, the researcher can either adopt the full procedure or, if desired, opt for using only part of the methodology (e.g., employ a predefined number of instances, use a predefined number of repetitions, force balanced runs on each instance, etc.).

\begin{algorithm}
	\caption{Full procedure for the comparison of algorithms}
	\label{alg:caiser}
	\begin{algorithmic}[1]
		\Require Set of available instances $\Gamma_S$; algorithms $a_1,a_2$; accuracy threshold $se^*$; initial sample size $n_0$; maximum sample size $n_{max}$; desired significance level $\alpha$, type-II error rate $\beta^*$, and MRES $d^*$.
		\State Calculate $N^*$ \Comment{Using \eqref{eq:t.samplesize1} or \eqref{eq:t.samplesize2}}
		\If{$N^*>\left|\Gamma_S\right|$}
			\State Investigate power, change parameters if needed. \Comment{Sec. \ref{sec:fixedSS}}
		\EndIf
		\State $\mbf{x} \leftarrow [\ ]$
		\For{$\min\left(N^*,\left|\Gamma_S\right|\right)$ times}
		\State Sample (without replacement) instance $\gamma_j\in\Gamma_S$
		\State Sample $a_1,a_2$ on $\gamma_j$ \Comment{Algorithm \ref{alg:reps_abs}}
		\State Calculate $\widehat{\phi}_j$ \Comment{Using \eqref{eq:diff_ab} or \eqref{eq:diff_perc}}
		\State $\mbf{x}\leftarrow\left[\mbf{x},\widehat{\phi}_j\right]$
		\EndFor
		\State Test of hypotheses using $\mbf{x}$ as the test sample
		\State Verify test assumptions \Comment{Secs. \ref{sec:assumptions}-\ref{sec:nonpar}} 
		\vspace{.10cm}
		\State\Return Test results; power profile (if needed)\;
	\end{algorithmic}
\end{algorithm}

Finally, as mentioned in Section \ref{sec:algcomp}, there are two sources of variability that affect the total variance $\sigma^2_{\Phi}$, namely the across-instances variance $\sigma^2_{\phi}$, which represents the variance of the values of paired differences in performance if all $\phi_j$ were precisely known; and the within-instances variance $\sigma^2_{\epsilon} = se^2_{\widehat{\phi}_j}$, which quantifies the ``measurement error'' on the values of $\phi_j$. Considering this, the standardized effect size used in the preceding discussion can be expressed as
\begin{equation}
\label{eq:cohend_2}
d = \frac{\delta}{\sigma_{\Phi}} = \frac{\delta}{\sqrt{\sigma^2_{\phi} + \sigma^2_{\epsilon}}}.
\end{equation}

While the experimenter can do little to change the across-instances variance, it is possible to reduce the standard error of estimation, as presented in Section \ref{sec:repetitions}. This composition of the total variance can be helpful in defining $se^*$ when calculating of the number of repetitions. Some guidelines are provided in Section \ref{sec:defpars}.

\subsection{Independence and normality}
\label{sec:assumptions}
The techniques presented so far have been based on two explicit assumptions: independence, i.e., the assumption that observations used for calculating the statistics of interest do not present any unmodeled dependencies, or that one observation does not influence another \cite{Montgomery2013,Sheskin2011}; and normality of the sampling distribution of the means.

In the case of this work, the assumption of independence can be guaranteed by design. In Algorithm \ref{alg:reps_abs}, the samples generated for the two algorithms on any given instance are produced without one observation influencing the value of any other -- e.g., by the usual (and rather obvious) practice of using different random seeds for different runs of randomized algorithms; or of using distinct, preferably randomly distributed initial points for deterministic methods. As for the paired test and sample size calculations, the assumption of independence can also be guaranteed by design. By using the values of $\widehat{\phi}_j$ as the individual observations we avoid the most common error in this kind of experiment, namely that of pseudoreplication \cite{Hurlbert1984,Lazic2010}, i.e., the use of repeated measurements $x_{ij}$ as if they were independent replicates. Ensuring independent algorithmic runs, as mentioned previously, also helps guarantee this assumption.


The other assumption, i.e., that of normality of the sampling distribution of the means, cannot be so easily guaranteed by design. It can, however, be verified \textit{a posteriori} without much effort. A first test of this assumption relies on the fact that, if the distribution of the data is normal, then the sampling distribution of the means will also be normal, regardless of the sample size. This suggests that a first test of normality can be performed on the data itself -- i.e., on the sets of observations $x_{1jk}$ and $x_{2jk}$ used to estimate $\phi_j$; and on the set of estimates $\widehat{\phi}_j$ used for testing the hypotheses of interest. Common statistical tests of normality include the Kolmogorov-Smirnov or Shapiro-Wilk tests \cite{Sheskin2011}, although in most cases visual inspection using a normal Q-Q plot is considered sufficient \cite{Montgomery2013}. If the data is found not to be significantly deviant from normality then the methods presented in this section can be considered accurate.

If the data itself deviates significantly from normality, an approximate test can be performed on the estimated sampling distribution of the means instead, e.g., using bootstrap \cite{Davison1997}. A quick (albeit computationally intensive) procedure for assessing normality of the sampling distribution of the means is to generate a vector $\mbf{y}_B$ of resampling estimates of the mean using a bootstrap procedure and then visually inspecting this vector using a normal Q-Q plot.\footnote{Using inferential tests on $\mbf{y}_B$ is not good practice, as the number of resamples can be made arbitrarily large, which would artificially inflate the degrees-of-freedom of any such test.} This assessment strategy is summarized in Algorithm \ref{alg:bootsdm}, where \verb|SampleWithReplacement|$\left(\mbf{y},n~\mbox{times}\right)$ means to build a vector of $n$ observations sampled (with replacement) from $\mbf{y}$. If this estimated sampling distribution of the mean does not deviate from normality, then the assumption can be considered satisfied for the  methods presented in this section.
\begin{algorithm}
	\caption{Bootstrapping the sampling distribution of the mean}
	\label{alg:bootsdm}
	\begin{algorithmic}[1]
		\Require Sample vector $\mbf{y}$; number of bootstrap resamplings $R$.
		\State $\bar{\mbf{y}}_B\leftarrow\left[\ \right]$ \Comment{Initialize empty vector}\;
		\State $n\leftarrow$ $dim(\mbf{y})$ \Comment{Vector length}\;
		\For{($R$ times)}
		\State $\mbf{y}_T \leftarrow$ SampleWithReplacement($\mbf{y},n$ times)\;
		\State $\bar{y}_T\leftarrow~mean(\mbf{y}_T)$\;
		\State $\bar{\mbf{y}}_B\leftarrow\left[\bar{\mbf{y}}_B, \bar{y}_T\right]$ \Comment{Append $\bar{y}_T$ to $\bar{\mbf{y}}_B$}\;
		\EndFor
		\vspace{.10cm}
		\State\Return $\bar{\mbf{y}}_B$\;
	\end{algorithmic}
\end{algorithm}

Finally, if the assumption of normality is violated (or expected to be, in the design phase of the experiment), one must employ nonparametric methods instead. A brief discussion of these techniques is provided next.

\subsection{Nonparametric alternatives}
\label{sec:nonpar}
If the assumption of normality of the sampling distribution of the means cannot be guaranteed,\footnote{Notice that it is relatively common for the normality assumption to be violated in the original data, but valid under transformations such as \textit{log} or \textit{square root}. The topic of data transformations is, however, outside the scope of this manuscript.} different procedures should be employed. Some possibilities for estimating $\phi_j$ and for testing hypotheses regarding the expected performance difference between two algorithms are presented in this section.

\subsubsection{Nonparametric estimation of $se_{\phi_j}$ and of the number of repetitions}
\label{sec:nonpar:phi.j}
When the assumptions regarding the sampling distribution of $\widehat{\phi}_j$ are not true,
the estimates of the standard error calculated in Section \ref{sec:repetitions} may be incorrect (particularly for the case of percent differences). If that is the case, a bootstrap approach can be used to estimate $se_{\widehat{\phi}_j}$ and, consequently, the required number of repetitions.

To obtain a bootstrap estimation of $se_{\widehat{\phi}_j}$, recall the definition of standard error as the standard deviation of the sampling distribution of a given estimator. A bootstrap estimator of $se_{\widehat{\phi}_j}$ can be calculated using the routine shown in Algorithm \ref{alg:bootSE}, and the value returned can then be used directly into Algorithm \ref{alg:reps_abs}.
\begin{algorithm}
	\caption{Bootstrap estimation of $se_{\widehat{\phi}_j}$}
	\label{alg:bootSE}
	\begin{algorithmic}[1]
		\Require Sample vectors $\mbf{x}_{1j},\mbf{x}_{2j}$; number of bootstrap runs $R$.
		\State $\widehat{\mbs{\phi}}_j\leftarrow\left[\ \right]$\;
		\State $n_{1j}\leftarrow$ $dim(\mbf{x}_{1j})$\;
		\State $n_{2j}\leftarrow$ $dim(\mbf{x}_{2j})$\;
		\For{($R$ times)}
		\State $\mbf{x}_{1}^{b} \leftarrow$ SampleWithReplacement($\mbf{x}_{1j},n_{1j}$ times)\;
		\State $\mbf{x}_{2}^{b} \leftarrow$ SampleWithReplacement($\mbf{x}_{2j},n_{2j}$ times)\;
		\State Calculate $\widehat{\phi}_{j,r}$ using \eqref{eq:diff_ab} or \eqref{eq:diff_perc}\;
		\State $\widehat{\mbs{\phi}}_j\leftarrow\left[\widehat{\mbs{\phi}}_j,\widehat{\phi}_{j,r}\right]$\;
		\EndFor
		\vspace{.10cm}
		\State\Return sample standard deviation of $\widehat{\mbs{\phi}}_j$\;
	\end{algorithmic}
\end{algorithm}

Notice that, unlike in the parametric approach, the optimal ratio $r_{opt}$ used in Algorithm \ref{alg:reps_abs} to determine which algorithm should be sampled may not be optimal in the theoretical sense when using a bootstrap estimate of $se_{\widehat{\phi}_j}$. Nonetheless, there are two arguments that can be advanced for using it in this case: first, it will always result in more intensive sampling of the algorithm presenting the greatest variance, which makes sense from the perspective of reducing the standard error of the estimates of $\phi_j$. Second, since the sampling distribution of the means will become progressively closer to a normal distribution as the sample sizes are increased, the estimation of $r_{opt}$ will become increasingly better as more observations are collected, and thus the sample sizes yielded by Algorithm \ref{alg:reps_abs} should approach optimality as the sampling progresses.

Finally, it is important to highlight that the bootstrap procedure tends to be considerably more computationally intensive than the parametric one, due to the resampling procedures involved in its calculation. This difference, however, becomes less important when the run times of $a_1,a_2$ are longer, needing, e.g., seconds or minutes to complete.

\subsubsection{Nonparametric tests of hypotheses}
\label{sec:nonpar:N}
Common alternatives for the paired t-test include \textit{Wilcoxon's signed-ranks test}, which assumes independence and symmetry about the median\footnote{Although it is very common in the literature on the experimental comparison of algorithms to ignore the fact that Wilcoxon's signed-ranks test works under the assumption of symmetry.} of $P\left(\Phi\right)$ \cite{Sheskin2011}; and the \textit{binomial sign test}, which requires only the assumption of independence \cite{Sheskin2011}, at the cost of reduced power. Both can be used to test hypotheses regarding the \textit{median} of $P\left(\Phi\right)$ instead of the mean, which is another way to quantify the expected differences between two algorithms.

The determination of the number of instances for these cases can be done using an argument based on the \textit{asymptotic relative efficiency} (ARE) of these tests relative to the paired t-test. The ARE can be defined \cite{Montgomery2013} as ``\textit{the limiting ratio of the sample sizes necessary to obtain identical error probabilities for the two procedures.}''. In the specific case of the Wilcoxon test, we have that \cite{Montgomery2013} ``\textit{For normal populations, the ARE of the Wilcoxon signed-rank test relative to the t-test is approximately 0.95; For non-normal populations, the ARE is at least 0.86, and in many cases it will exceed unity.}". As for the binomial sign test, the ARE is $0.637$ \cite{Sheskin2011}, showing its more conservative characteristic.

Under these considerations, a reasonable rule-of-thumb is to calculate the required number of instances using the formulas for the paired t-test, and then dividing the value of $N^*$ by the ARE of the test under normality:
\begin{equation}
\begin{split}
N^*_{wilc} &= N^*/0.86 \approxeq 1.16 N^*\\
N^*_{sign} &= N^*/0.637 \approxeq 1.57 N^*\\
\end{split}
\end{equation}

Notice that i) these are conservative estimates, reflecting (particularly in the case of Wilcoxon's test) an expected worst-case scenario, which means that the actual power can be larger than the one used for the calculations; and ii) the binomial sign test requires over $50\%$ more instances to achieve the same power under this supposed worst-case scenario, which may be unreasonable in many situations. However, if $P\left(\Phi\right)$ is severely skewed, this may be the only test for which the assumptions can be maintained (i.e., for which nominal error rates can be reasonably expected to hold), and as such it remains as an interesting last resource.\footnote{There are other ways to calculate the sample size for the binomial sign test that are less conservative, but for the sake of brevity this will not be discussed here.}

\subsection{The case of predefined $N$}
\label{sec:fixedSS}
The second part of the proposed methodology, described in Sections \ref{sec:replicates} and \ref{sec:nonpar:N}, is concerned with estimating the smallest number of instances required for achieving predefined statistical properties for a given experiment. This estimation can be very useful in several distinct situations, e.g., when designing test sets for specific problem classes, or when performing experiments on algorithms for computationally expensive optimization problems \cite{Jones1998}. As mentioned earlier, however, there are cases in which it may not be possible to arbitrarily choose the sample size for a given experiment. Common examples include situations when only a limited number of instances is available, or when a predefined test set needs to be employed, as is often the case in standardized comparison experiments \cite{BBOB}. 

\begin{figure}[ht]
	\centering
	\includegraphics[width = 0.7\textwidth]{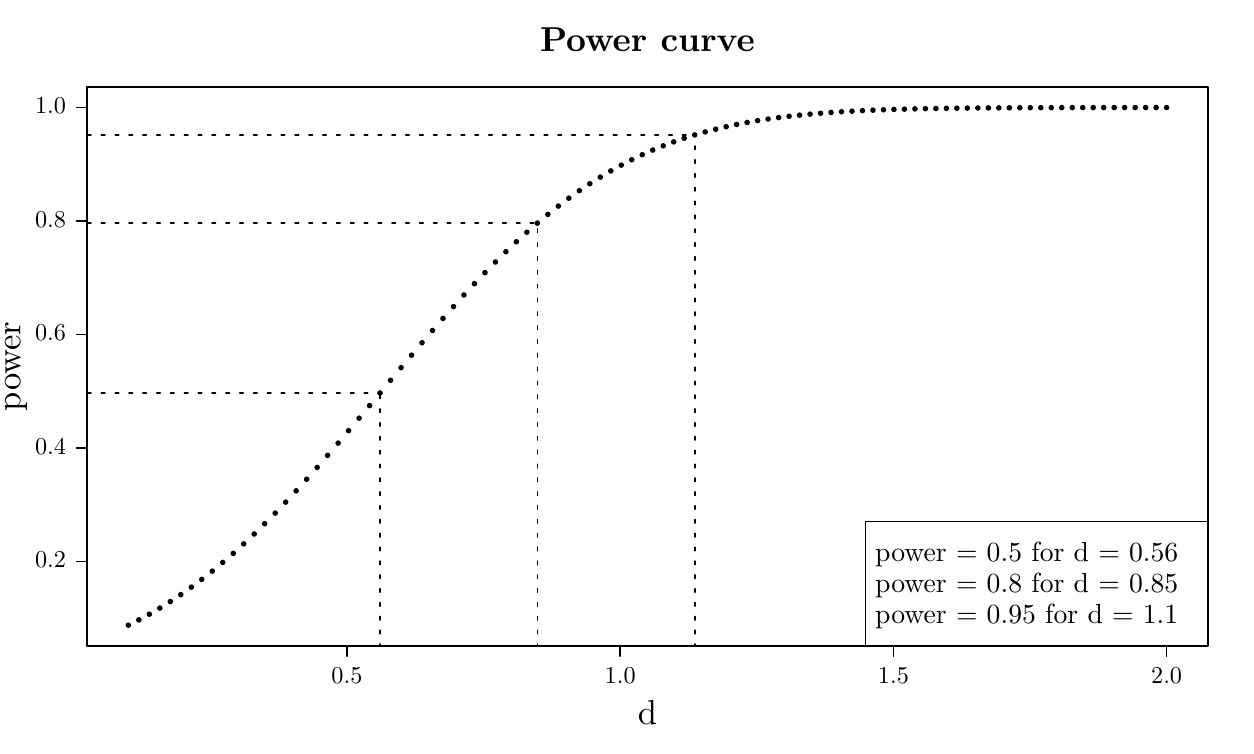}
	\caption{Example of power curve that can be derived in cases with a predefined number of instances. }
	\label{fig:powercurve}
\end{figure}

Even if that is the case, however, it is still possible to employ the principles discussed in the preceding sections to obtain a better perspective of the statistical properties of the experiment. For instance, for predefined $N$, the proposed methodology can still be used to determine the number of runs of each algorithm on each instance, so as to guarantee a desired standard error for the paired differences in each instance. Moreover, the sample size calculations provided in Section \ref{sec:replicates} can be easily adapted to maintain a fixed $N$ and estimate instead other relevant properties, e.g., the expected statistical power for a given MRES. For instance, by keeping $\alpha$ and $N$ fixed in \eqref{eq:t.power}, one can iterate over different values of $ncp = d/\sqrt{N}$ and obtain a power curve for a fixed-sample experiment, prior to actually collecting the data. Figure \ref{fig:powercurve} provides an example of this kind of power curve, which can be quite useful for researchers interested in evaluating which differences between algorithms could the experiment be reasonably expected to detect. Similar curves can be constructed for other pairs of power-related variables, e.g., maintaining a fixed power and iterating over $N$ to obtain a curve of effect sizes $d$ for which that power is expected as a function of sample size.

\subsection{Defining reasonable experimental parameters}
\label{sec:defpars}
Finally,  it is important to discuss the choice of reasonable values for the experimental parameters. In terms of the proposed methodology, the required parameters are shown at the beginning of Algorithm \ref{alg:caiser}. The set of available instances $\Gamma_S$ and the algorithms to be compared, $a_1,a_2$, are relatively straightforward - $\Gamma_S$ is usually a list of available instances, which may or may not be exhausted in the experiment, and $a_1,a_2$ are the algorithms to be compared.

The definition of the remaining user-defined parameters for the experimental protocol -- namely $se^*, n_0$ and $n_{max}$ for calculating the number of repetitions, and $\alpha,\beta^*$ and $d^*$ for the number of replicates -- is a little more subtle. Starting with the statistical error rates, $\alpha$ and $\beta^*$ should ideally be defined based on the consequences of the errors they control - i.e., the consequences of falsely detecting a nonexistent difference, or of failing to detect an existing one. However, defining these consequences can be very challenging even in experiments with more easily quantifiable  consequences, and in practice ``standard'' values are often used -- $0.05$ or $0.01$ for $\alpha$, and $0.2$ or $0.15$ for $\beta^*$. It is important to recall that (i) there is nothing inherently special about these values, they are simply conventions that can, and often should, be challenged; and (ii) there is a tradeoff between the error rates and the sample size, so that the lower these values, the larger the number of instances will be needed to control both errors at their nominal values for a given MRES.

Determining a good value for the MRES is also heavily experiment-dependent, since a small difference in one context could be considered substantial in another. In our discussions we have been using the standardized effect size $d$ for the power calculations, in which case the MRES, $d^*$, should be selected based on units of standard deviations - e.g., a $d^* = 0.5$ would mean that we are interested in detecting differences equal to or larger than one half standard deviation. While some fields possess somewhat standard target values for ``small'', ``medium'' and ``large'' effects (see, e.g., the discussion by \cite{Sawilowsky2009}), researchers should be aware that specific features of different application areas can and should take precedence over application-agnostic predefined values.

When $\phi_j$ is defined as the percent differences (Sec. \ref{sec:perc_diff}), it may be more intuitive to use the simple effect size, $\delta$, instead of the standardized one. This would allow statements such as ``we are interested in detecting mean performance gains of more than $5\%$'', which tend to be more straightforward. In this case, however, a reasonable upper bound for the total standard deviation -- the denominator of the r.h.s.~of \eqref{eq:cohend_2} -- must be provided by the user. Such bound may be obtained using either a pilot study, estimated from published results, or defined using previous knowledge about the algorithms tested.

Regarding the experimental parameters necessary for estimating the number of runs on each instance, $n_{0}$ should ideally be set based on the expected shape of the distribution of observations of algorithm performance -- bell-shaped distributions can use lower $n_0$ (values as low as $3$ or $4$ are sufficient for the sampling distribution of means to converge to a Gaussian shape in these cases), other symmetric distributions can use intermediate values (e.g., $10$), and more strongly skewed distributions should use larger values ($n_0 = 20$ or $30$). If the distribution is severely skewed, it is often more practical to work on log-transformed data, which tends to bring the distribution to a more well-behaved shape \cite{Crawley2013}. The value of $n_{max}$ should be selected based on the available computational budget for the experiment, but knowing that lower values will result in sample sizes that may fail to control the within-instances error $se_{\widehat{\phi}_j}$ at the predefined level $se^*$, which can result in reduced overall power for the experiment.

The definition of the measurement error threshold, $se^*$, should be performed in such a way that this component of the total standard deviation does not dominate the power calculations -- in other words, the value of $se^*$ should be much smaller than the expected across-instances variance -- e.g., $\left(se^*\right)^2 \leq 0.1\sigma_{\Phi}^2$.

Finally, it is important to remember that even if the number of available instances is much larger than the calculated $N^*$ and the researchers desire (or are required) to employ all in the comparison, the methodology presented in this section can still provide precious information - both for the determination of the number of runs on each instance and, critically, for defining a MRES prior to the experiment, so that the results obtained are interpreted under the light of practical relevance, and not only statistical significance.

\section{Examples of application}
\label{sec:examples}
In this section we present results aimed at demonstrating the application of the proposed method for the definition of sample sizes in the experimental comparison of two algorithms. Since the objective in this section is to illustrate the comparison methodology, and not to generate new results regarding the performance of specific algorithms, we opted for using algorithms for which an existing implementation is readily available. The implementation of the proposed methodology for calculating the sample sizes is available as the \texttt{R} package \texttt{CAISEr} (\textit{Comparison of Algorithms with Iterative Sample-size Estimation in R}) \cite{caiser}, which is also introduced in the Supplemental Materials.

\subsection{Experiment 1}
For the first experiment we assume a situation in which we wish to compare two versions of the \textit{Multiobjective Evolutionary Algorithm Based on Decomposition} (MOEA/D) \cite{Zhang2007,Li2009,Campelo2017}, in terms of their mean IGD \cite{Zitzler2003}, for which smaller values indicate better performance. . The first version is the one presented in Section V-E  of the original MOEA/D paper \cite{Zhang2007}, and the second is a modified MOEA/D known as MOEA/D-DE, proposed by Li and Zhang in 2009 \cite{Li2009}. The specific parameters of these two algorithms are summarized in Table \ref{tab:moead-algs}, and a detailed explanation can be found in the relevant literature \cite{Campelo2017,Zhang2007,Li2009}. 

\renewcommand{\arraystretch}{1.3}
\begin{table}[!htb]
	\centering
	\footnotesize
	\caption{Algorithms and parameters. \textbf{Boldface} entries highlight differences. $n_v$ denotes the dimension of the problem instance being solved.}
	\begin{tabular}{c|c|c}
		\hline
		Component & Alg. 1: MOEA/D & Alg. 2: MOEA/D-DE\\
		\hline
		Decomposition strategy								& SLD ($H = 99$) & SLD ($H = 99$)\\
		\hline
		Neighborhood strategy& By weight vectors ($T = 20, \delta = 1.0$)& By weight vectors ($T = 20, \mbs{\delta =} \mbf{0.9}$)\\
		\hline
		Aggregation function									& Weighted Tchebycheff		& Weighted Tchebycheff\\
		\hline
		\multirow{3}{*}{Variation Operators}	& \multirow{2}{*}{SBX ($\eta = 20, p_c = 1$)}& \textbf{Differential mutation} \verb|/rand/1| ($F = 0.5$)\\
		&																				& \textbf{Binomial recombination} ($CR = 1.0$)\\
		& Polynomial mutation ($\eta = 20, p_m = 1/n_v$) & Polynomial mutation ($\eta = 20, p_m = 1/n_v$) \\
		\hline
		Update strategy											& Standard update													  &	\textbf{Restricted update }($n_r = 2$)\\
		\hline
		Stop criterion											& $2000n_v$ function calls													  &	$2000n_v$ function calls\\
		\hline
	\end{tabular}
	\label{tab:moead-algs}
\end{table}

Suppose that we wish to compare the performance of these two algorithms on a hypothetical problem class based on the \textit{UF} benchmark set \cite{Zhang2008}, defined as \textit{the set of all possible problems for which functions \textit{UF1} to \textit{UF7}, with dimensions $n_v\in\left[10,40\right]$, can be considered representative}. While in this case one would be justifiably interested in using all available test instances (a total of $217$) to obtain a more complete understanding of the behavior of these algorithms on the problem class of interest, the resulting computational cost of such an exhaustive experiment may be quite large.\footnote{While in this particular example the required computational budget for exhausting all available instances would not be unattainable, limitations to the number of instances that can be reasonably employed in an experiment can be much more severe when researching, for instance, heuristics for optimizing numerical models in engineering applications, or other expensive optimization scenarios \cite{Tenne2010}. The present example was inspired in part by the authors' past experience with such problems.} Consequently, a first step in comparing these two methods may be to investigate whether they present differences in mean IGD performance that exceed some minimal threshold of practical relevance, which can be achieved using a subset of the available instances, at a computational cost much smaller than what would be required for the full investigation. 

To this end we used the proposed methodology to investigate the mean \textit{percent} differences of performance between the two algorithms summarized in Table \ref{tab:moead-algs} on the problem class of interest. The parameters used for this experiment were defined as follows: $\alpha = 0.05,~\beta^* = 0.2,~d^* = 0.5,~n_0 = 15,~n_{max} = 200,~se^* = 0.05$. The standard errors were calculated using the bootstrap approach (Algorithm \ref{alg:bootSE}), using $R = 999$; and the number of instances was calculated assuming the use of a t test (Section \ref{sec:replicates}) and a bilateral alternative hypothesis \eqref{eq:t.samplesize1}.\footnote{The full replication script for this experiment is available in the Vignette ``\textit{Adapting Algorithms for CAISEr}'' of the \texttt{CAISEr} package \cite{caiser}.}

Following the procedure outlined in Algorithm \ref{alg:caiser}, the proposed methodology indicated that the required number of instances in this case was $N^* = 34$. This amount of instances was randomly sampled (without replacement) from the set of available instances, and the two algorithms were run on each instance according to the procedure defined in Algorithm \ref{alg:reps_abs}. The results of this process are summarized in Table \ref{tab:res1}.

\renewcommand{\arraystretch}{1.3}
\begin{table*}[!htb]
	\centering
	\caption{Summary of results obtained in Experiment 1. Instances marked in \textbf{boldface} were sampled up to the maximum allowed budget, $n_{max} = 200$.}
	\begin{tabular}{c|c|c|c|c?c|c|c|c|c}
 \hline
 Instance (dim.) &$\widehat{\phi}_j$ & $\widehat{se}_{\widehat{\phi}_j}$&$n_{1j}$ & $n_{2j}$&Instance (dim.) & $\widehat{\phi}_j$ & $\widehat{se}_{\widehat{\phi}_j}$&$n_{1j}$ & $n_{2j}$\\
 \hline
 UF4 (13) & -0.14 & 0.02 & 15 & 15 & \textbf{UF5 (17)} & 0.46 & 0.05 & 83 & 117 \\ 
 UF2 (29) & -0.36 & 0.05 & 65 & 15 & UF3 (15) & -0.08 & 0.05 & 40 & 53 \\ 
 \textbf{UF5 (28)} & 0.69 & 0.05 & 80 & 120 & UF4 (16) & -0.11 & 0.03 & 15 & 15 \\ 
 UF1 (29) & -0.63 & 0.05 & 25 & 15 & UF7 (18) & -0.89 & 0.05 & 41 & 33 \\ 
 UF2 (36) & -0.29 & 0.05 & 71 & 16 & UF7 (38) & -0.86 & 0.05 & 32 & 16 \\ 
\textbf{UF3 (29)} & 0.05 & 0.05 & 99 & 101 & UF4 (14) & -0.21 & 0.02 & 15 & 15 \\ 
 UF3 (10) & 0.07 & 0.05 & 57 & 58 & UF1 (11) & -0.82 & 0.02 & 15 & 15 \\ 
 UF7 (16) & -0.95 & 0.00 & 15 & 15 & UF1 (16) & -0.75 & 0.02 & 15 & 15 \\ 
 UF7 (29) & -0.90 & 0.04 & 15 & 15 & UF2 (32) & -0.35 & 0.05 & 51 & 15 \\ 
 UF2 (25) & -0.38 & 0.05 & 66 & 15 & UF3 (24) & -0.19 & 0.05 & 42 & 47 \\
  UF4 (30) & -0.04 & 0.02 & 15 & 15 & UF6 (34) & -0.74 & 0.05 & 15 & 15 \\ 
  UF1 (26) & -0.65 & 0.05 & 15 & 15 & UF4 (32) & -0.00 & 0.03 & 15 & 15 \\ 
  UF2 (18) & -0.46 & 0.05 & 40 & 15 & UF2 (11) & -0.47 & 0.05 & 46 & 15 \\ 
  UF7 (36) & -0.92 & 0.02 & 15 & 15 & UF2 (22) & -0.54 & 0.05 & 44 & 15 \\ 
  UF4 (18) & -0.17 & 0.02 & 15 & 15 & UF1 (17) & -0.71 & 0.03 & 15 & 15 \\ 
  UF2 (34) & -0.40 & 0.05 & 44 & 15 & UF1 (18) & -0.69 & 0.03 & 15 & 15 \\ 
  UF2 (39) & -0.29 & 0.05 & 71 & 15 & UF3 (23) & -0.18 & 0.05 & 33 & 58 \\
 \hline
	\end{tabular}
	\label{tab:res1}
\end{table*}

Some interesting remarks can be made regarding the results summarized in Table \ref{tab:res1}. First, 
we observed negative values of $\widehat{\phi}_j$ in the majority of instances tested, 
suggesting an advantage of the MOEA/D-DE over the original MOEA/D (recall that smaller IGD is better). MOEA/D-DE also seems to require less repetitions in most instances, which indicates lower variance on several instances, a desirable feature since it means that the algorithm tends to return more consistent performance values across repeated runs. 

Another noteworthy point is that in three of the 34 instances sampled -- UF3 (29), UF5 (17), and UF5 (28), boldfaced in the table -- the maximum allocated budget ($n_{max} = 200$) was not enough to reduce the standard error $\widehat{se}_{\widehat{\phi}_j}$ below the predefined threshold of $se^* = 0.05$. In these three cases the second algorithm, MOEA/D-DE, seems to present an unusually high variance (evidenced by the large number of runs attributed to it by the proposed sampling methodology), resulting in the need for a larger number of repeated runs to reduce the uncertainty on the estimate of $\widehat{\phi}_j$. However, since the resulting standard errors in these three cases were not particularly high\footnote{More specifically: $\widehat{se}_{\widehat{\phi}_j} = 0.0518$ for \textit{UF5 (28)}; $\widehat{se}_{\widehat{\phi}_j} = 0.0544$ for \textit{UF3 (29); and $\widehat{se}_{\widehat{\phi}_j} = 0.0536$ for \textit{UF5 (17)}}}, their effect on the total residual variance is likely negligible.

Continuing with the experimental procedure outlined in algorithm \ref{alg:caiser}, a t-test performed on our sample of estimated paired differences of performance yields statistically significant results ($p = 2.90\times10^{-6}, df=33$) with an estimated paired mean difference in IGD of $\widehat{\mu}_D = -0.379$ ($CI_{0.95} = [-0.517, -0.242]$), which means an expected value of IGD for the MOEA/D-DE that is $\left(37.9\pm13.7\right)\%$ better than that of the original MOEA/D for our problem class of interest.

\begin{figure}[ht]
	\centering
	\includegraphics[width = 0.6\textwidth, trim={0 .5cm 0 .7cm},clip]{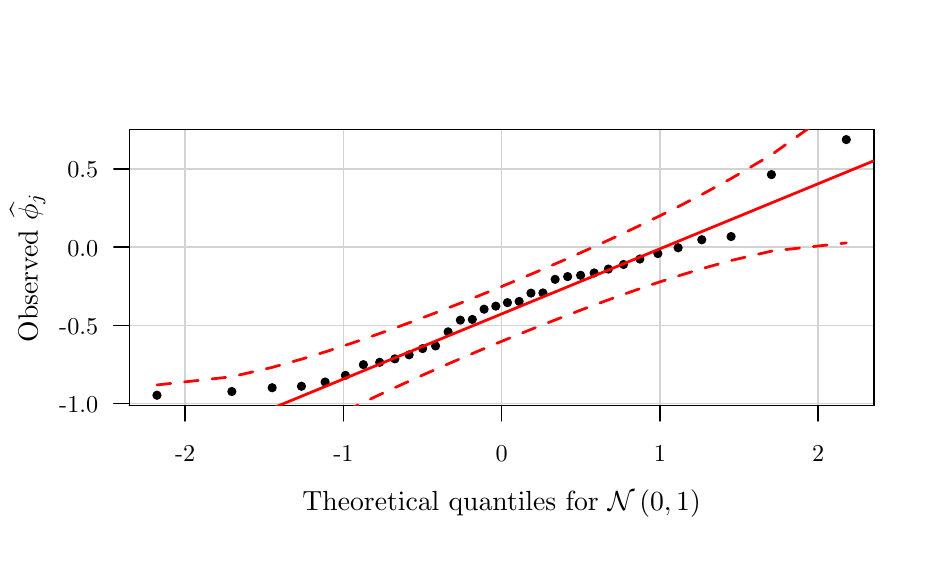}
	\caption{Normal quantile-quantile plot for observations $\widehat{\phi}_j$ in Experiment 1.}
	\label{fig:qqplot1}
\end{figure}

The normality assumption of the t-test can be easily validated using the normal QQ-plot shown in Figure \ref{fig:qqplot1}. The plot indicates that no expressive deviations of normality are present, which gives us confidence in using the t test as our inferential procedure of choice, since the sampling distribution of the means will be even closer to a Normal variable than the data distribution, diluting whatever small deviations from normality may be present.

\begin{figure}[htb]
	\centering
	\includegraphics[width = 0.7\textwidth]{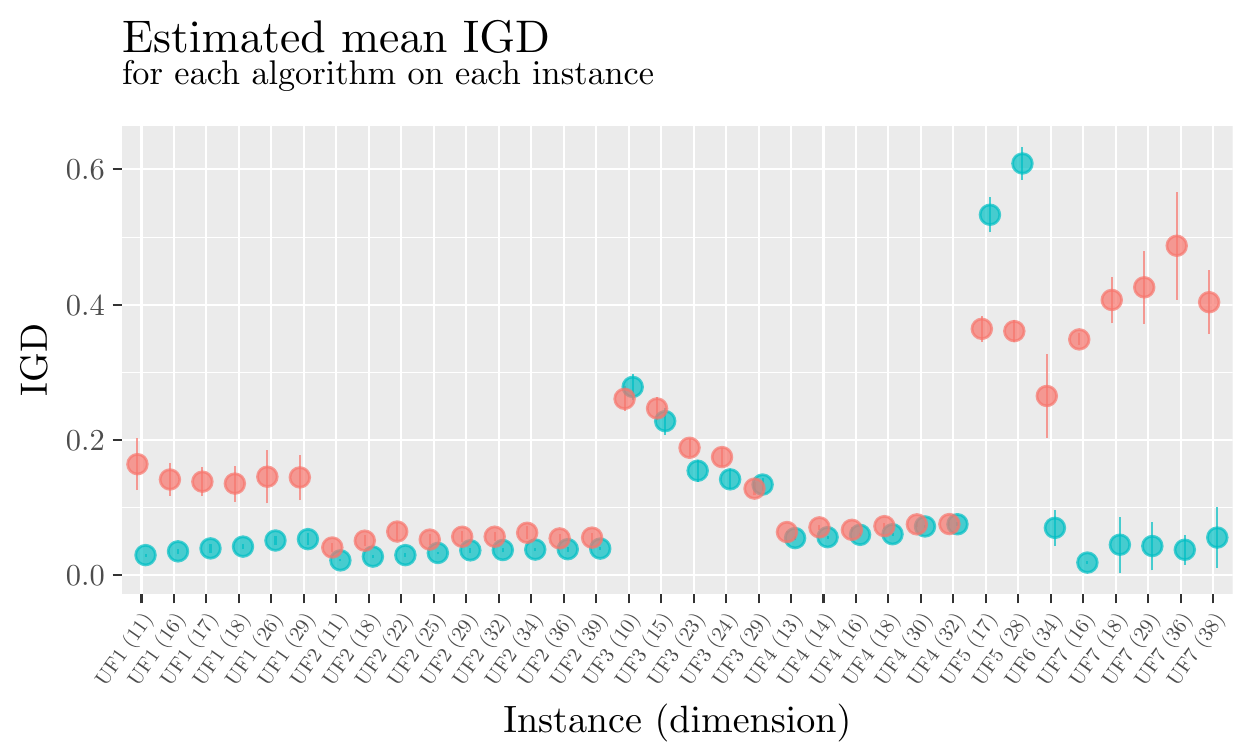}
	\includegraphics[width = 0.85\textwidth]{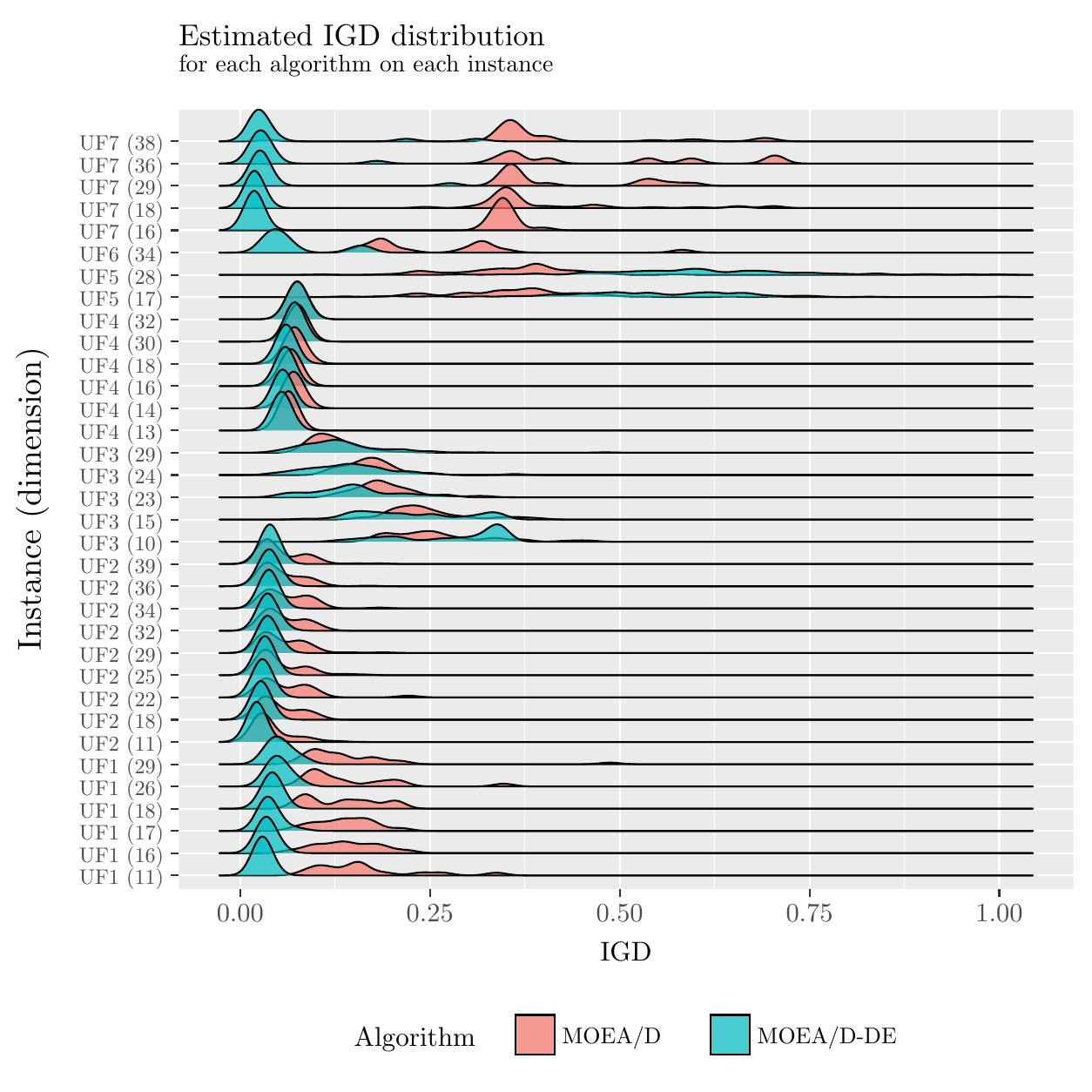}
	\caption{\textbf{Top}: $95\%$ confidence intervals on the means of individual IGD values of each algorithm on each sampled instance. \textbf{Bottom}: Density estimates of IGD for MOEA/D and MOEA/D-DE on each sampled instance. Notice the discrepant performance of MOEA/D-DE on instance \textit{UF5}. }
	\label{fig:results1}
\end{figure}

Finally, it is important to reinforce that these results could also be used to motivate further analyses of the performance of these two algorithms for problems belonging to the problem class of interest, even before proceeding to the full, exhaustive test on all available instance. For example, the individual IGD distributions and mean values of each algorithm on each instance, presented in Figure \ref{fig:results1}, suggest that both algorithms encounter difficulties when solving \textit{UF5} (and, to a lesser extent, \textit{UF3}) instances, which could motivate a more focused investigation into the reasons for these poor performance profiles, and on possible algorithmic improvements to remedy this problem. A natural follow-up to the experiment presented in this first example would be to broaden the investigation to include the full available test set, in which case the proposed methodology could still be useful in defining the number of repetitions to be performed for each algorithm on each test instance, as well as the expected statistical power of whatever subgroup comparisons the researcher could deem interesting.

\subsection{Experiment 2}
As mentioned in Section \ref{sec:fixedSS}, the proposed methodology can also be useful in situations when the researcher uses a predefined set of benchmark instances to compare two algorithms. To illustrate this case, we used a set of $200$ large instances of the unrelated parallel machines problem with sequence dependent setup times, provided by Vallada and Ruiz \cite{Vallada2011} for calibration experiments.\footnote{The instance files can be retrieved from \url{http://soa.iti.es/problem-instances}} Currently the best results for this problem are those presented by Santos \textit{et al.} \cite{Santos2016} using a simulated annealing algorithm with six neighborhood structures (\textit{Shift, Switch, Task move, Swap, Two-Shift,} and \textit{Direct swap}), randomly selected at each trial move.\footnote{The source codes used for this experiment can be retrieved from \url{http://github.com/andremaravilha/upmsp-scheduling}}.

Preliminary tests suggest that the most influential neighborhood structure for this case is \textit{Task move}, which presents the largest expected improvement value across a wide range of problem sizes. To isolate and quantify the effect of this specific neighborhood structure to the performance of the method, two versions of the algorithm were compared: a \textit{full version}, which is the original algorithm equipped with all six neighborhood structures; and a \textit{no-task-move} version, which uses exactly the same structure but does not include the \textit{Task move} neighborhood. As mentioned above, these two versions were tested on the calibration test set proposed by Vallada and Ruiz \cite{Vallada2011}, which features $200$ large instances with $M\in\left\{10,15,20,25,30\right\}$ machines and $N\in\left\{50,100,150,200,250\right\}$ jobs. All algorithmic aspects were set exactly as in \cite{Santos2016}, with the stop criteria employed at each instance being the total run time, calculated as a function of instance size following guidelines from the original references \cite{Vallada2011,Santos2016}.

\clearpage

\begin{figure}[!h]
	\centering
	\includegraphics[width = 0.6\textwidth]{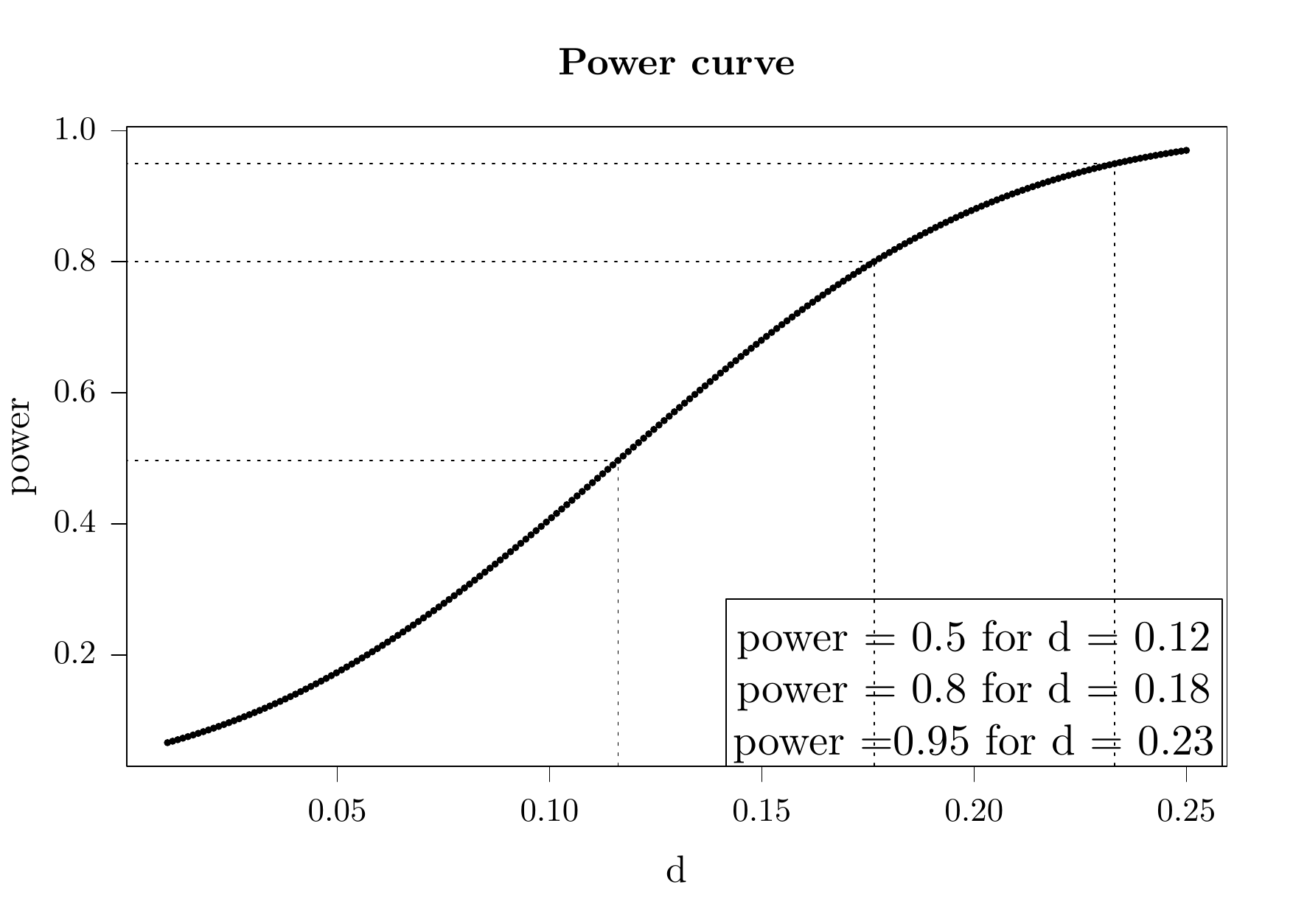}
	\caption{Expected sensitivity of experiment 2 to different effect sizes, for a t-test test with a one-sided alternative hypothesis. With $200$ instances we can be fairly confident that the experiment will be able to identify mean performance gains greater than approximately 0.2 standard deviations.}
	\label{fig:powercurve2}
\end{figure}

Given that the number of instances is predefined, there is no need to calculate it using the approach presented in Section \ref{sec:replicates}. Instead, we used the proposed methodology to estimate the power curve of the experiment, that is, the expected sensitivity of this comparison to detect effects of different magnitudes. This is illustrated in Figure \ref{fig:powercurve2}, which was derived assuming that the desired significance of the experiment is $\alpha = 0.05$, and that a t-test test will be performed using a one-sided alternative hypothesis, since we have a prior expectation that the \textit{full version} algorithm should be better than the \textit{no-task-move}, and are interested in testing and quantifying this effect. 

As suggested in the figure, this experiment has a reasonable probability of detecting mean performance gains due to the use of the \textit{Task move} neighborhood structure greater than about $0.2$ standard deviations. Smaller differences in mean performance, particularly under about $0.1$ standard deviations, can go undetected, but in terms of impact on the expected behavior of the algorithm these would be really minor effects.

The experiment was performed using the proposed method for iteratively estimating the required number of repetitions for each algorithm on each of the $200$ instances. The experimental parameters were set as $se^* = 0.05$ on the percent differences, $n_0 = 15$ and $n_{max} = 150$. The standard errors were calculated using the parametric formulas provided in Section \ref{sec:perc_diff}. A t test performed on the resulting data suggested significant differences at the $95\%$ confidence level ($p < 2\times 10^{-16}, df = 199$, against a one-sided, lower $H_1$) with an estimated paired mean difference of $\widehat{\mu}_D = -0.361$ ($CI_{0.95} = \left[-0.380, -0.342\right]$), which means that the expected impact of the \textit{Task move} neighborhood on the performance of the algorithm, for an instance belonging to the same problem family defined by the test, set is a reduction of $\left(36.1\pm 1.9 \right)\%$ in the makespan of the final solution returned.\footnote{The graphical analysis of the residuals did not suggest expressive deviations of normality. The results table and residual analysis are provided in the Supplemental Materials.} 

\begin{figure}[!h]
	\centering
	\includegraphics[width = \textwidth]{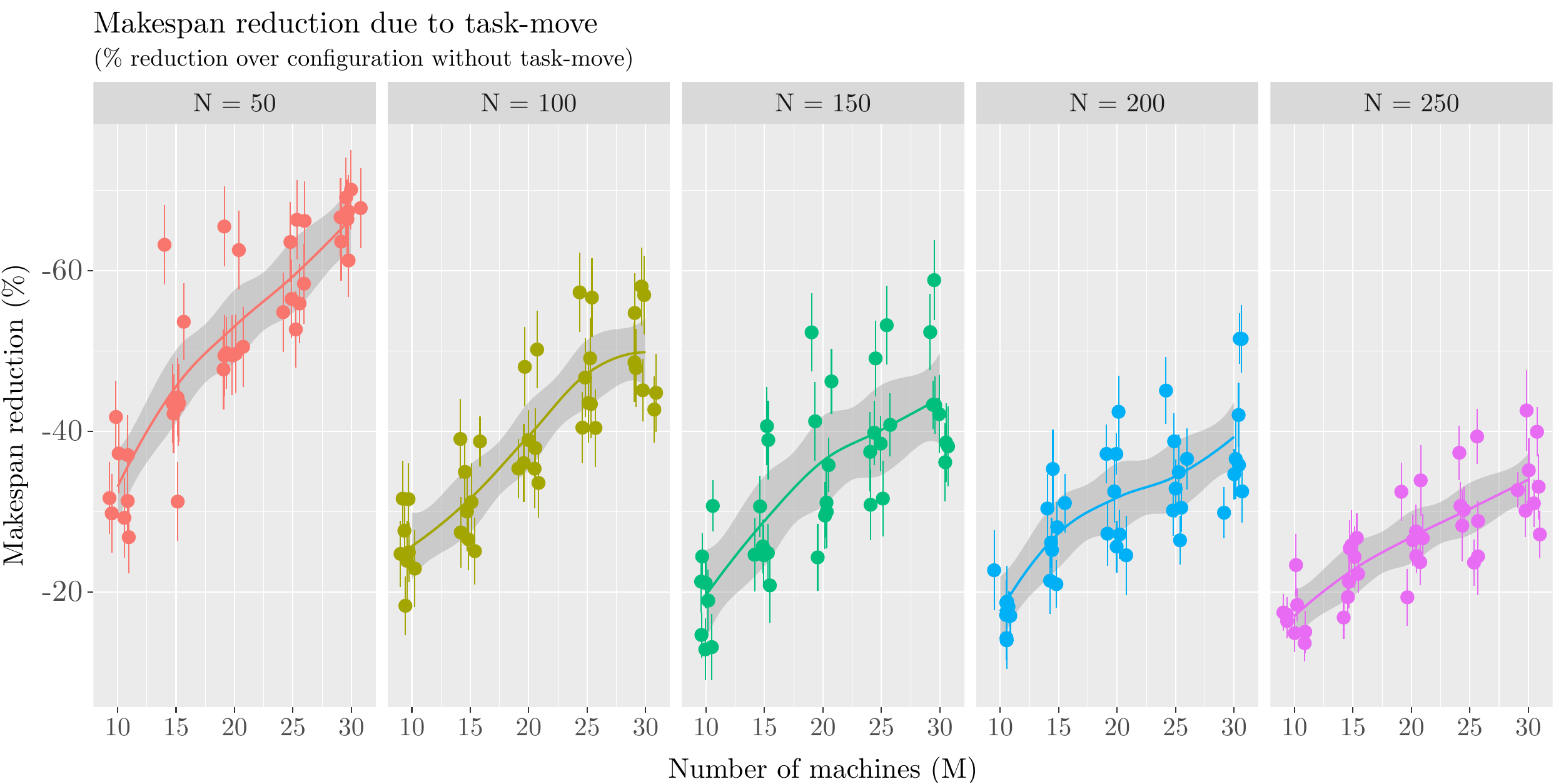}
	\caption{Percent gains in performance attributable to using the \textit{Task-move} neighborhood. The y-axis indicates how much lower the average makespan was for the full algorithm in comparison to the \textit{no-task-move} version (notice that the y-axis is vertically reversed). Vertical lines represent the standard errors of each observation. The x-coordinates of the observations were perturbed slightly around their true values ($M = \left\{10,15,20,25,30\right\}$), for visualization purposes.}
	\label{fig:TM_effect}
\end{figure}

Notice that further analyses could (and should) be performed on this same data, to refine the conclusions and, possibly, suggest new lines of inquiry. For instance, while the overall expected improvement due to the use of \textit{Task move} in the pool of possible movements is quantified as $\left(36.1\pm 1.9 \right)\%$, knowledge, e.g., of instance size can improve the estimation accuracy of performance gains. This is illustrated in Figure \ref{fig:TM_effect}, which suggests that, while the use of \textit{Task move} provides relevant improvements across all problem sizes tested, its effect increases with the number of machines ($M$) and decreases with the number of jobs ($N$). A detailed quantification of these effects and the reasons behind them is, however, outside the scope of the present work.

\vspace{-1em}

\clearpage
\section{Conclusions}
\label{sec:conclusions}
Experimental comparisons play a central role in the research and development of optimization heuristics. In this work we proposed a methodology to address an often neglected aspect of the design of experiments for the comparison of two algorithms on a given problem class, namely the determination of adequate sample sizes, in terms of the amount of problem instances to be employed in a given comparison as well as the number of repeated runs of each algorithm on each instance. 

Prior to describing the methodology for sample size calculations, it was important to formally define the \textit{algorithm comparison problem} considered in our work. This definition, presented in Section \ref{sec:algcomp}, was important for two reasons: first, it allowed us to delimit the scope of the scientific and statistical questions of interest in this particular paper, and to formalize the population which our statistical procedures would attempt to interrogate. Second, we hope that it may provide statistical grounds for future discussions and developments in the field of experimental comparisons of algorithms.

Based on the concepts defined in Sections \ref{sec:algcomp} and \ref{sec:stats}, the proposed methodology for calculating the relevant sample sizes for the comparison of two algorithms on a given problem class was presented in Section \ref{sec:caiser}. The determination of the number of instances is based on considerations of \textit{practically relevant differences} and on the desired statistical power to detect them. Analytic formulas were presented for the parametric case, using both one-sided and two-sided alternative hypotheses. Nonparametric approximations based on the asymptotic relative efficiency of the Wilcoxon signed-ranks and the binomial sign test were also provided.

The number of runs of each algorithm on each instance is determined iteratively, approximating optimal sample size ratios for the reduction of the uncertainty associated with the estimation of paired differences in performance. Analytic solutions for the optimal ratio of sample sizes were provided for the simple and percent difference cases, based on the assumption of normality of the sampling distribution of the means. These results were also useful as approximations for the calculation of the standard errors using bootstrap, for cases in which the assumption of normality described in Section \ref{sec:assumptions} cannot be expected or guaranteed.

Examples of application were provided in Section \ref{sec:examples}, illustrating the potential of the proposed methodology to provide researchers in the field with a methodologically sound, reproducible way of determining the required numbers of instances and runs, as well as to identify limitations with existing experimental benchmark sets. Two common situations were discussed: the first, in which the researcher needs to determine how many instances of a given problem class to use for a given test, as well as how many runs each algorithm should perform on each instance; and a second one, which focused on the use of the proposed methodology for an experiment dealing with a fixed-size, predefined benchmark set, in which case the assessment of the sensitivity of the experiment to different effect sizes replaces the estimation of the required number of instances to achieve a redefined statistical power.

It is important to reinforce that the proposed methodology is by no means an universal way to test algorithms: when the goal of the experiment is to characterize an heuristic, how robust it is and its best / worst case performance behavior, different methodologies can and should be employed. However, such extensive experimentation is prohibitive in a number of scenarios, such as in several cases of applied engineering optimization \cite{Tenne2010} or when comparing heuristics on very large, time-consuming instances.

One of the main aims of this work was to lay the statistical and methodological groundwork for the calculation of required sample sizes in the experimental comparison of algorithms. While the developments and results presented do fulfill this particular goal, there are a number of limitations and possibilities of continuity that can be explored. We finish this work by examining a few of the most promising ones.

\subsection{Limitations and Possibilities}
Possibly the main limitations of the methodology developed in this paper are, in order of severity: (i) the fact that it is only defined for the comparison of two algorithms; (ii) the fact that the definition of the number of instances is performed \textit{a priori}, using a fixed sample size methodology; and (iii) the fact that only centrality statistics (the mean and, to a certain extent, the median) were considered. Below we offer a brief discussion these three points, and offer our views of what can be done to further extend the proposed methodology.

Regarding the number of algorithms considered in the comparison, a natural next step of this work is to extend the sample size estimation methodologies for multiple algorithms. This can be achieved in a relatively straightforward manner for the estimation of the number of instances, using standard formulas for either omnibus tests (e.g., ANOVA, Friedman) or planning directly for the eventual post-hoc pairwise comparisons \cite{Mathews2010}. Estimating the number of runs, however, will require greater improvements on the method proposed in this work, probably based on the definition of standard error thresholds for each individual algorithm on each instance, instead of on the standard error of the differences.

While the \textit{a priori} definition of the number of instances provides a reasonable expectation of statistical power for a given MRES, the required sample size may be considerably smaller if the actual effect size is much larger than the predefined $d^*$. Using sequential analysis methodologies, such as the ones commonly employed in clinical trials or industrial settings \cite{Botella2006,Bartroff2013}, may result in a reduced number of instances being necessary to determine the existence of differences between two (or eventually more) algorithms, and represent another possibility of continuity for this work. In this aspect, Bayesian alternatives to the comparison of algorithms can be of particular interest, since they may allow the aggregation of existing knowledge in the form of prior probability distributions, as well as the incremental aggregation of observations without the need for cumbersome significance corrections \cite{Kruschke2010}.

The possibility of using the methodology defined in this work as a framework for comparisons of algorithms according to different statistics -- e.g., variances, rates of convergence, regression coefficients, or best/worst cases -- is yet another promising direction. While most experiments still focus on average (mean/median) cases, the need for methodologically sound comparisons of other quantities has long been recognized \cite{Hooker1996,Eiben2002}, and we believe the methodology presented in this paper can be easily adapted for such comparisons. First, the bootstrap approach for the calculation of the number of runs can be extended to different measures of paired differences in performance - medians, quantiles, or other statistics - in a relatively straightforward manner (using balanced samples if needed, or deriving optimal ratios for these statistics). Moreover, standard statistical tests for other quantities tend to be readily available, as well as analytic formulas for power and sample size \cite{Mathews2010}, providing a rich basis upon which better, more comprehensive protocols for algorithmic comparisons can be built.


\begin{thebibliography}{10}
	\providecommand{\url}[1]{{#1}}
	\providecommand{\urlprefix}{URL }
	\expandafter\ifx\csname urlstyle\endcsname\relax
	\providecommand{\doi}[1]{DOI~\discretionary{}{}{}#1}\else
	\providecommand{\doi}{DOI~\discretionary{}{}{}\begingroup
		\urlstyle{rm}\Url}\fi
	
	\bibitem{Amo2012}
	del Amo, I.G., Pelta, D.A., Gonz{\'{a}}lez, J.R., Masegosa, A.D.: An algorithm
	comparison for dynamic optimization problems.
	\newblock Applied Soft Computing \textbf{12}(10), 3176--3192 (2012).
	
	\bibitem{Barr1995}
	Barr, R.S., Golden, B.L., Kelly, J.P., Resende, M.G.C., Stewart, W.R.:
	Designing and reporting on computational experiments with heuristic methods.
	\newblock Journal of Heuristics \textbf{1}(1), 9--32 (1995).
	
	\bibitem{Bartroff2013}
	Bartroff, J., Lai, T., Shih, M.C.: Sequential Experimentation in Clinical
	Trials: Design and Analysis.
	\newblock Springer (2013)
	
	\bibitem{BartzBeielstein2005}
	Bartz-Beielstein, T.: {New Experimentalism Applied to Evolutionary
		Computation}.
	\newblock Ph.D. thesis, Universit\"at Dortmund, Germany (2005)
	
	\bibitem{Bartz-Beielstein2006}
	Bartz-Beielstein, T.: Experimental Research in Evolutionary Computation.
	\newblock Springer (2006)
	
	\bibitem{Bartz-Beielstein2010}
	Bartz-Beielstein, T., Chiarandini, M., Paquete, L., Preuss, M.: Experimental
	Methods for the Analysis of Optimization Algorithms.
	\newblock Springer (2010)
	
	\bibitem{Bausell2006}
	Bausell, R., Li, Y.F.: Power analysis for experimental research: a practical
	guide for the biological, medical and social sciences.
	\newblock Cambridge University Press (2006)
	
	\bibitem{Benavoli2014}
	Benavoli, A., Corani, G., Mangili, F., Zaffalon, M., Ruggeri, F.: A bayesian
	wilcoxon signed-rank test based on the dirichlet process.
	\newblock In: 30th International conference on machine learning, pp. 1026--1034
	(2014)
	
	\bibitem{Birattari2004}
	Birattari, M.: On the estimation of the expected performance of a metaheuristic
	on a class of instances: how many instances, how many runs?
	\newblock Tech. Rep. IRIDIA/2004-001, Université Libre de Bruxelles, Belgium
	(2004)
	
	\bibitem{Birattari2009}
	Birattari, M.: Tuning Metaheuristics -- A Machine Learning Perspective.
	\newblock Springer Berlin Heidelberg (2009)
	
	\bibitem{Birattari2007}
	Birattari, M., Dorigo, M.: {How to assess and report the performance of a
		stochastic algorithm on a benchmark problem: Mean or best result on a number
		of runs?}
	\newblock Optimization Letters \textbf{1}, 309--311 (2007)
	
	\bibitem{Botella2006}
	Botella, J., Xim{\'{e}}nez, C., Revuelta, J., Suero, M.: Optimization of sample
	size in controlled experiments: The {CLAST} rule.
	\newblock Behavior Research Methods \textbf{38}(1), 65--76 (2006).
	
	\bibitem{Efron1994:book}
	Bradley~Efron, R.T.: An Introduction to the Bootstrap, 1 edn.
	\newblock Chapman and Hall (1994)
	
	\bibitem{Campelo2017}
	Campelo, F., Batista, L.S., Aranha, C.: The {MOEADr} package -- a
	component-based framework for multiobjective evolutionary algorithms based on
	decomposition.
	\newblock Submitted: Journal of Statistical Software  (2017)
	
	\bibitem{caiser}
	Campelo, F., Takahashi, F.: {CAISEr: Comparison of Algorithms with Iterative
		Sample Size Estimation} (2017).
	\newblock \urlprefix\url{https://CRAN.R-project.org/package=CAISEr}
	
	\bibitem{Carrano2011}
	Carrano, E.G., Wanner, E.F., Takahashi, R.H.C.: A multicriteria statistical
	based comparison methodology for evaluating evolutionary algorithms.
	\newblock {IEEE} Transactions on Evolutionary Computation \textbf{15}(6),
	848--870 (2011).
	
	\bibitem{Chow2003}
	Chow, S.C., Wang, H., Shao, J.: Sample Size Calculations in Clinical Research.
	\newblock CRC Press (2003)
	
	\bibitem{CoffinSaltzman2000}
	Coffin, M., Saltzman, M.J.: Statistical analysis of computational tests of
	algorithms and heuristics.
	\newblock {INFORMS} Journal on Computing \textbf{12}(1), 24--44 (2000).
	
	\bibitem{Crawley2013}
	Crawley, M.: The {R} Book, 2nd. edn.
	\newblock Wiley (2013)
	
	\bibitem{Czarn2004}
	Czarn, A., MacNish, C., Vijayan, K., Turlach, B.: Statistical exploratory
	analysis of genetic algorithms: the importance of interaction.
	\newblock In: Proceedings of the 2004 {IEEE} Congress on Evolutionary
	Computation. Institute of Electrical {\&} Electronics Engineers ({IEEE})
	(2004).
	
	\bibitem{Davison1997}
	Davison, A.C., Hinkley, D.V.: Bootstrap methods and their application.
	\newblock Cambridge University Press (1997)
	
	\bibitem{Demsar2006}
	Dem\v{s}ar, J.: Statistical comparisons of classifiers over multiple data sets.
	\newblock Journal of Machine Learning Research \textbf{7}, 1--30 (2006)
	
	\bibitem{Derrac2014}
	Derrac, J., Garc{\'{i}}a, S., Hui, S., Suganthan, P.N., Herrera, F.: Analyzing
	convergence performance of evolutionary algorithms: A statistical approach.
	\newblock Information Sciences \textbf{289}, 41--58 (2014)
	
	\bibitem{Derrac2011}
	Derrac, J., Garc{\'{\i}}a, S., Molina, D., Herrera, F.: A practical tutorial on
	the use of nonparametric statistical tests as a methodology for comparing
	evolutionary and swarm intelligence algorithms.
	\newblock Swarm and Evolutionary Computation \textbf{1}(1), 3--18 (2011).
	
	\bibitem{Eiben2002}
	Eiben, A., Jelasity, M.: A critical note on experimental research methodology
	in {EC}.
	\newblock In: Proceedings of the 2002 {IEEE}Congress on Evolutionary
	Computation. Institute of Electrical {\&} Electronics Engineers ({IEEE})
	(2002).
	
	\bibitem{Fieller1954}
	Fieller, E.C.: Some problems in interval estimation.
	\newblock Journal of the Royal Statistical Society. Series B (Methodological)
	\textbf{16}(2), 175--185 (1954)
	
	\bibitem{Franz2007}
	Franz, V.: Ratios: A short guide to confidence limits and proper use (2007).
	\newblock \urlprefix\url{https://arxiv.org/pdf/0710.2024v1.pdf}
	
	\bibitem{Garcia2009}
	Garc{\'{\i}}a, S., Fern{\'{a}}ndez, A., Luengo, J., Herrera, F.: A study of
	statistical techniques and performance measures for genetics-based machine
	learning: accuracy and interpretability.
	\newblock Soft Computing \textbf{13}(10), 959--977 (2009).
	
	\bibitem{Garcia2010}
	Garc{\'{\i}}a, S., Fern{\'{a}}ndez, A., Luengo, J., Herrera, F.: Advanced
	nonparametric tests for multiple comparisons in the design of experiments in
	computational intelligence and data mining: Experimental analysis of power.
	\newblock Information Sciences \textbf{180}(10), 2044--2064 (2010).
	
	\bibitem{Garcia2008}
	Garc{\'{\i}}a, S., Molina, D., Lozano, M., Herrera, F.: A study on the use of
	non-parametric tests for analyzing the evolutionary algorithms' behaviour: a
	case study on the {CEC}'2005 {S}pecial session on real parameter
	optimization.
	\newblock Journal of Heuristics \textbf{15}(6), 617--644 (2008).
	
	\bibitem{Grissom2012}
	Grissom, R.J., Kim, J.J.: Effect Sizes for Research, 2nd edn.
	\newblock Routledge (2012)
	
	\bibitem{BBOB}
	Hansen, N., Auger, A., Mersmann, O., Tusar, T., Brockhoff, D.: {COCO:} {A}
	platform for comparing continuous optimizers in a black-box setting.
	\newblock CoRR \textbf{abs/1603.08785} (2016).
	\newblock \urlprefix\url{http://arxiv.org/abs/1603.08785}
	
	\bibitem{COCO2016}
	Hansen, N., T\v{u}sar, T., Mersmann, O., Auger, A., Brockoff, D.: {COCO}: The
	experimental procedure (2016).
	\newblock \urlprefix\url{https://arxiv.org/abs/1603.08776}
	
	\bibitem{Hooker1996}
	Hooker, J.N.: Testing heuristics: We have it all wrong.
	\newblock Journal of Heuristics \textbf{1}(1), 33--42 (1996)
	
	\bibitem{Hurlbert1984}
	Hurlbert, S.H.: Pseudoreplication and the design of ecological field
	experiments.
	\newblock Ecological Monographs \textbf{54}(2), 187--211 (1984).
	
	\bibitem{Jain1991}
	Jain, R.K.: The Art of Computer Systems Performance Analysis.
	\newblock John Wiley and Sons Ltd (1991)
	
	\bibitem{Johnson2002}
	Johnson, D.: A theoretician's guide to the experimental analysis of algorithms.
	\newblock In: M.~Goldwasser, D.~Johnson, C.~McGeoch (eds.) Data Structures,
	Near Neighbor Searches, and Methodology: Fifth and Sixth DIMACS
	Implementation Challenges, \emph{DIMACS Series in Discrete Mathematics and
		Theoretical Computer Science}, vol.~59, pp. 215--250. American Mathematical
	Society (2002)
	
	\bibitem{Jones1998}
	Jones, D.R., Schonlau, M., Welch, W.J.: Efficient global optimization of
	expensive black-box functions.
	\newblock Journal of Global Optimization \textbf{13}(4), 455--492 (1998)
	
	\bibitem{Krohling2015}
	Krohling, R.A., Lourenzutti, R., Campos, M.: Ranking and comparing evolutionary
	algorithms with hellinger-{TOPSIS}.
	\newblock Applied Soft Computing \textbf{37}, 217--226 (2015).
	
	\bibitem{Kruschke2010}
	Kruschke, J.K.: Doing Bayesian Data Analysis: A Tutorial with R and BUGS, 1st
	edn.
	\newblock Academic Press, Inc. (2010)
	
	\bibitem{Lazic2010}
	Lazic, S.E.: The problem of pseudoreplication in neuroscientific studies: is it
	affecting your analysis?
	\newblock {BMC} Neuroscience \textbf{11}(5), 397--407 (2010).
	
	\bibitem{Lenth2001}
	Lenth, R.V.: Some practical guidelines for effective sample size determination.
	\newblock The American Statistician \textbf{55}(3), 187--193 (2001)
	
	\bibitem{Li2009}
	Li, H., Zhang, Q.: Multiobjective optimization problems with complicated pareto
	sets, {MOEA}/d and {NSGA}-{II}.
	\newblock {IEEE} Transactions on Evolutionary Computation \textbf{13}(2),
	284--302 (2009).
	
	\bibitem{Maravilha2019}
	Maravilha, A.L., Pereira, L.M., Campelo, F.: Statistical characterization of
	neighborhood structures for the unrelated parallel machine problem with
	sequence-dependent setup times.
	\newblock In preparation
	
	\bibitem{Mathews2010}
	Mathews, P.: Sample Size Calculations: Practical Methods for Engineers and
	Scientists, 1st edn.
	\newblock Matthews Malnar \& Bailey Inc. (2010)
	
	\bibitem{McGeoch1996}
	McGeoch, C.C.: Feature article{\textemdash}toward an experimental method for
	algorithm simulation.
	\newblock {INFORMS} Journal on Computing \textbf{8}(1), 1--15 (1996).
	
	\bibitem{Montgomery2013}
	Montgomery, D.C., Runger, G.C.: Applied Statistics and Probability for
	Engineers, 6th edn.
	\newblock Wiley (2013)
	
	\bibitem{Mori2015}
	Mori, T., Sato, Y., Adriano, R., Igarashi, H.: Optimal design of {RF} energy
	harvesting device using genetic algorithm.
	\newblock Sensing and Imaging \textbf{16}(1) (2015)
	
	\bibitem{Nuzzo2014}
	Nuzzo, R.: Scientific method: Statistical errors.
	\newblock Nature \textbf{506}(7487), 150--152 (2014)
	
	\bibitem{R}
	{R Core Team}: R: A Language and Environment for Statistical Computing.
	\newblock R Foundation for Statistical Computing, Vienna, Austria (2017).
	\newblock \urlprefix\url{https://www.R-project.org/}
	
	\bibitem{Ridge2007}
	Ridge, E.: {Design of Experiments for the Tuning of Optimisation Algorithms}.
	\newblock Ph.D. thesis, The University of York, UK (2007)
	
	\bibitem{Santos2016}
	Santos, H.G., Toffolo, T.A., Silva, C.L., Berghe, G.V.: Analysis of stochastic
	local search methods for the unrelated parallel machine scheduling problem.
	\newblock International Transactions in Operational Research \textbf{00}
	(2016).
	\newblock \urlprefix\url{https://doi.org/10.1111/itor.12316}
	
	\bibitem{Sawilowsky2009}
	Sawilowsky, S.S.: New effect size rules of thumb.
	\newblock Journal of Modern Applied Statistical Methods \textbf{8}(2), 597--599
	(2009)
	
	\bibitem{Shaffer1995}
	Shaffer, J.P.: Multiple hypothesis testing.
	\newblock Annual review of psychology \textbf{46}(1), 561--584 (1995)
	
	\bibitem{Sheskin2011}
	Sheskin, D.J.: Handbook of Parametric and Nonparametric Statistical Procedures.
	\newblock Taylor \& Francis (2011)
	
	\bibitem{Tenne2010}
	Tenne, Y., Goh, C.K.: Computational Intelligence in Expensive Optimization
	Problems.
	\newblock Springer (2010)
	
	\bibitem{Vallada2011}
	Vallada, E., Ruiz, R.: A genetic algorithm for the unrelated parallel machine
	scheduling problem with sequence dependent setup times.
	\newblock European Journal of Operational Research \textbf{211}(3), 612--622
	(2011)
	
	\bibitem{Yuan2004}
	Yuan, B., Gallagher, M.: {Statistical racing techniques for improved empirical
		evaluation of evolutionary algorithms}.
	\newblock Parallel Problem Solving From Nature - {PPSN VIII} \textbf{3242},
	172--181 (2004)
	
	\bibitem{YuanGallagher2009}
	Yuan, B., Gallagher, M.: An improved small-sample statistical test for
	comparing the success rates of evolutionary algorithms.
	\newblock In: Proceedings of the 11th Annual conference on Genetic and
	evolutionary computation - {GECCO}09. Association for Computing Machinery
	({ACM}) (2009).
	
	\bibitem{Zhang2007}
	Zhang, Q., Li, H.: {MOEA}/d: A multiobjective evolutionary algorithm based on
	decomposition.
	\newblock {IEEE} Transactions on Evolutionary Computation \textbf{11}(6),
	712--731 (2007).
	
	\bibitem{Zhang2008}
	Zhang, Q., Zhou, A., Zhao, S., Suganthan, P., Liu, W., Tiwari, S.:
	Multiobjective optimization test instances for the cec 2009 special session
	and competition.
	\newblock Tech. Rep. CES-887, University of Essex (2008).
	\newblock
	\urlprefix\url{http://dces.essex.ac.uk/staff/zhang/moeacompetition09.htm}.
	\newblock (revised on 20/04/2009)
	
	\bibitem{Zitzler2003}
	Zitzler, E., Thiele, L., Laumanns, M., Fonseca, C., {Fonseca}, V.: {Performance
		assessment of multiobjective optimizers: An analysis and review}.
	\newblock IEEE Transactions on Evolutionary Computation \textbf{7}(2), 117--132
	(2003)
	
\end{thebibliography}
\end{document}


\maketitle

\section{Detailed Derivation of the number of Repetitions}

\subsection{Using the Simple Difference of Two Means}
\label{sec:abs_diff}
Assume that we are interested in using the simple difference of mean
performance between algorithms \(a_1,a_2\) on each instance as our
values of \(\phi_j\). 
In this case we define  
\(\phi_j = \mu_{2j} - \mu_{1j}\), for which the sample estimator is
given by
\begin{equation}
\label{eq:diff_ab}
\widehat{\phi}_j^{(1)} = \widehat{\Delta\mu} = \widehat{\mu}_{2j} - \widehat{\mu}_{1j} = \bar{X}_{2j} - \bar{X}_{1j},
\end{equation}

\noindent where \(\bar{X}_{ij}\) is the sample mean of algorithm \(a_i\)
on instance \(\gamma_j\). Let the distribution of performance values of
algorithm \(a_i\) on instance \(\gamma_j\) be expressed as an (unknown)
probability density function with expected value \(\mu_{ij}\) and
variance \(\sigma^2_{ij}\), i.e., \[
x_{ij} \sim X_{ij} = \mathcal{P}\left(\mu_{ij}, \sigma^2_{ij}\right).
\]

Assuming that the conditions of the Central Limit Theorem hold, we
expect
\(\bar{X}_{ij} \sim \mathcal{N}\left(\mu_{ij},\sigma^2_{ij}/n_{ij}\right)\)
and, consequently,
\begin{equation}
\label{eq:D_abs}
\widehat{\phi}_j^{(1)}\sim \mathcal{N}\left(\mu_{2j} - \mu_{1j},\frac{\sigma^2_{1j}}{n_{1j}} + \frac{\sigma^2_{2j}}{n_{2j}}\right).
\end{equation}

By definition, the standard error of \(\widehat{\phi}_j^{(1)}\) is the
standard deviation of this sampling distribution of the estimator, \[
se_{\widehat{\phi}_j^{(1)}} = \sqrt{\sigma^2_{1j}n_{1j}^{-1} + \sigma^2_{2j}n_{2j}^{-1}}.
\]

Given a desired upper limit for the standard error, \(se^*\), the
optimal sample sizes for the two algorithms \(a1,a2\) on instance
\(\gamma_j\) can be obtained by solving the optimization problem defined
as
\begin{equation}
\label{eq:opt_repsize}
\begin{split}
\mbox{Minimize: }& f(\mathbf{n}_j)  = n_{1j} + n_{2j},\\
\mbox{Subject to: }& g(\mathbf{n}_j) = se_{\widehat{\phi}_j^{(1)}} - se^* \leq 0.
\end{split}
\end{equation}

This problem can be solved analytically\footnote{Actually we solve a relaxed version of the problem which admits non-integer $n_{ij}$.} using the Karush-Kuhn-Tucker
(KKT) optimality conditions,
\begin{equation}
\label{eq:opt_sys}
\begin{aligned}
\nabla f(\mathbf{n}_j)  + \beta\nabla g(\mathbf{n}_j) &= 0,\\
\beta g(\mathbf{n}_j) &= 0,\\
\beta&\geq 0.
\end{aligned}
\end{equation}

The partial derivatives of the objective and constraint functions are
\begin{equation}
\label{eq:parc_der_abs}
\begin{split}
\frac{\partial f}{\partial n_{1j}} &= \frac{\partial f}{\partial n_{2j}} = 1,\\
\frac{\partial g}{\partial n_{1j}} &= -\frac{\sigma_{1j}^2}{n_{1j}^2\sqrt{\sigma^2_{1j}n_{1j}^{-1} + \sigma^2_{2j}n_{2j}^{-1}}},\\
\frac{\partial g}{\partial n_{2j}} &= -\frac{\sigma_{2j}^2}{n_{2j}^2\sqrt{\sigma^2_{1j}n_{1j}^{-1} + \sigma^2_{2j}n_{2j}^{-1}}}.
\end{split}
\end{equation}

Substituting the derivatives in \eqref{eq:parc_der_abs} into
\eqref{eq:opt_sys} and simplifying the terms results in the system:
\begin{align}
\beta\sigma_{1j}^2n_{1j}^{-2} &= 1,\label{eq:18}\\
\beta\sigma_{2j}^2n_{2j}^{-2} &= 1,\label{eq:19}\\
\sigma_{1j}^2n_{1j}^{-1} + \sigma_{2j}^2n_{2j}^{-1} &= \left(se^*\right)^2,\label{eq:20}
\end{align}

\noindent with \(\beta > 0\) (simple inspection of
\eqref{eq:18} shows \(\beta=0\) to be an impossibility). Equating the l.h.s. of \eqref{eq:18} with that of \eqref{eq:19} and rearranging the terms yields the optimal ratio of
sample sizes,
\begin{equation}
\label{eq:opt_ratio}
r_{opt} = \frac{n_{1j}}{n_{2j}} = \frac{\sigma_{1j}}{\sigma_{2j}},
\end{equation}

\noindent which means that algorithms \(a_1\) and \(a_2\) must be
sampled on instance \(\gamma_j\) in direct proportion to the standard
deviations of their performances on that instance. The result in
\eqref{eq:opt_ratio} is known in the statistical literature \cite{Mathews2010} as the \emph{optimal allocation of resources} for the estimation
of confidence intervals on the simple difference of two means. 

%

Since the populational variances $\sigma_{1j}^2,\sigma_{2j}^2$ are usually unknown, their values need to be estimated from the data. This results in the sample standard error,
\begin{equation}
\label{eq:se_abs}
\widehat{se}_{\widehat{\phi}_j^{(1)}} = \sqrt{s^2_{1j}n_{1j}^{-1} + s^2_{2j}n_{2j}^{-1}}.
\end{equation}

A good approximation of the optimal ratio of sample sizes can be similarly obtained by replacing $\sigma_{ij}$ by $s_{ij}$ in
\eqref{eq:opt_ratio}. This requires that an initial sample size \(n_0\) be obtained for each algorithm, to calculate initial estimates of \(s_{1j},s_{2j}\)\footnote{The definition of an
	initial value of \(n_0\) also helps increasing the probability that the
	sampling distributions of the means will be
	approximately normal.}, suggesting the iterative procedure described in Algorithm \ref{alg:reps_abs}, where \verb|Sample|$\left(a_i,\gamma_j,n~\mbox{times}\right)$ means to obtain $n$ observations of algorithm $a_i$ on instance $\gamma_j$.  To prevent an explosion of the number of repetitions in the case of poorly specified threshold values $se^*$ or particularly high-variance algorithms, a maximum budget $n_{max}$ can also be defined for the sampling on a given problem instance.
\begin{algorithm}
	\caption{Sample algorithms on one instance.}
	\label{alg:reps_abs}
	\begin{algorithmic}[1]
		\Require Instance $\gamma_j$; Algorithms $a_1,a_2$; accuracy threshold $se^*$; initial sample size $n_0$; maximum sample size $n_{max}$.
		\State $\mbf{x}_{1j}\leftarrow$ Sample$\left(a_1,\gamma_j,n_0~\mbox{times}\right)$
		\State $\mbf{x}_{2j}\leftarrow$ Sample$\left(a_2,\gamma_j,n_0~\mbox{times}\right)$
		\State $n_{1j}\leftarrow n_0$\;
		\State $n_{2j}\leftarrow n_0$\;
		\State Calculate $\widehat{se}$\;
		\While{$\left(\widehat{se} > se^*\right)$ \& $\left(n_{1j}+n_{2j}< n_{max}\right)$}
		\State Calculate $r_{opt}$\;
		\If{$\left(n_{1j}/n_{2j} < r_{opt}\right)$}
		\State $x\leftarrow$ Sample$\left(a_1,\gamma_j,1~\mbox{time}\right)$\;
		\State $\mbf{x}_{1j}\leftarrow [\mbf{x}_{1j}, x]$\;
		\Else
		\State $x\leftarrow$ Sample$\left(a_2,\gamma_j,1~\mbox{time}\right)$\;
		\State $\mbf{x}_{2j}\leftarrow [\mbf{x}_{2j}, x]$\;
		\EndIf
		\State Calculate $\widehat{se}$\;
		\EndWhile
		\vspace{.10cm}
		\State\Return $\mbf{x}_{1j},~\mbf{x}_{2j}$\;
	\end{algorithmic}
\end{algorithm}

After performing the procedure shown in Algorithm \ref{alg:reps_abs}, the estimate $\widehat{\phi}_j$ can be calculated using the vectors of observations $\mbf{x}_{1j}$ and $\mbf{x}_{2j}$.

\subsection{Using the Percent Difference of Two Means}
\label{sec:perc_diff}
While the approach of defining \(\phi_j\) as the simple difference
between the means of algorithms \(a_1,a_2\) on a given instance
\(\gamma_j\) is certainly useful, it may be subject to some difficulties. In
particular, defining a single precision threshold \(se^*\) for problem classes
containing instances with vastly different scales can be problematic and
lead to wasteful sampling. In these cases, it is generally more
practical and more intuitive to define the differences in performance
\(\phi_j\) as the \textit{percent mean gains} of algorithm \(a_2\) over
\(a_1\). In
this case we define\footnote{Considering a comparison where larger is better.}
\(\phi_j = \left(\mu_{2j} - \mu_{1j}\right)/\mu_{1j}\), for which the
sample estimator is 
\begin{equation}
\label{eq:diff_perc}
\widehat{\phi}_j^{(2)} = \widehat{\Delta\mu}_{(\%)} = \frac{\bar{X}_{2j} - \bar{X}_{1j}}{\bar{X}_{1j}} = \frac{\widehat{\phi}_j^{(1)}}{\bar{X}_{1j}}
\end{equation}

For this definition to be used we need to consider an
additional assumption, namely that
\(P(\bar{X}_{1j}\leq 0) \rightarrow 0\) (which is guaranteed, for
instance, when objective function values are always strictly positive,
which is common in several problem
classes).\footnote{If this assumption cannot be guaranteed, the use of percent differences is not advisable, and the researcher should instead perform comparisons using the simple differences.}
The distribution of \(\phi_j^{(1)}\) is given in \eqref{eq:D_abs}, which
means that under our working assumptions \(\widehat{\phi}_j^{(2)}\) is
distributed as the ratio of two independent normal
variables.\footnote{The independence between $\phi_j^{(1)}$ and $\bar{X}_{1j}$ is guaranteed as long as $X_{1j}$ and $X_{2j}$ are independent.}

A commonly used estimator of the standard error of
\(\widehat{\phi}_j^{(2)}\) is based on confidence interval derivations
by Fieller \cite{Fieller1954}. Considering the assumption that
\(P(\bar{X}_{1j}\leq 0) \rightarrow 0\), a simplified form of Fieller's
estimator can be used \cite{Franz2007}, which provides good coverage
properties. Under balanced sampling, i.e., \(n_{1j} = n_{2j} = n_j\),
the standard error is given as
\begin{equation}
\label{eq:FiellerSE}
\widehat{se}_{\widehat{\phi}_j^{(2)}} = \left|\widehat{\phi}_j^{(2)}\right| \left[ \frac{s^2_{1j}/n_j}{\bar{x}^2_{1j}} + \frac{\left(s^2_{1j}/n_j + s^2_{2j}/n_j\right)}{\left(\widehat{\phi}_j^{(1)}\right)^2}+ \frac{2}{n_j}\frac{cov\left(\mathbf{x}_{1j},\left(\mathbf{x}_{2j}-\mathbf{x}_{1j}\right)\right)}{\widehat{\phi}_j^{(1)}\bar{x}_{1j}} \right]^{1 / 2},
\end{equation}

\noindent where \(\mathbf{x}_{ij}\in \mathbb{R}^{n_j}\) represents the
vector of observations of algorithm \(a_i\) on instance \(\gamma_j\);
and \(cov\left(\cdot,\cdot\right)\) is the sample covariance of two
vectors.

Under the assumption of within-instance independence, i.e., that
\(X_{1j}\) and \(X_{2j}\) are independent, the expected value of
covariance will be close to zero, allowing us to disregard the covariance
term in \eqref{eq:FiellerSE}. This offers two advantages: first, it
simplifies calculations of the standard error, particularly for larger
sample sizes. Second, and more importantly, it allows us to consider
unbalanced sampling, as we did for the case of simple differences, 
which can lead to gains in efficiency. Removing
the covariance term, replacing the \(n_j\) dividing each sample standard
deviation by the corresponding \(n_{ij}\) and simplifying
\eqref{eq:FiellerSE} results in
\begin{equation}
\label{eq:FiellerSE_nocov}
\widehat{se}_{\widehat{\phi}_j^{(2)}} = \left|\widehat{\phi}_j^{(2)}\right|\sqrt{c_1n_{1j}^{-1} + c_2n_{2j}^{-1}},
\end{equation}

\noindent with
\begin{equation}
\label{eq:SE_nocov_c}
\begin{split}
c_1 &= s^2_{1j}\left[\left(\widehat{\phi}_j^{(1)}\right)^{-2} +\left(\bar{x}_{1j}\right)^{-2}\right];\\
c_2 &= s^2_{2j}\left(\widehat{\phi}_j^{(1)}\right)^{-2}.
\end{split}
\end{equation}

The problem of calculating the smallest total sample size required to
achieve a desired accuracy is equivalent to the one stated in
\eqref{eq:opt_repsize} (substituting
\(\widehat{se}_{\widehat{\phi}_j^{(1)}}\) by
\(\widehat{se}_{\widehat{\phi}_j^{(2)}}\) in the constraint function) and can be solved in a similar manner to yield the optimal ratio of sample sizes in the case of percent differences. 

The partial derivatives of \(f\left(\mathbf{n}_j\right)\) and \(g\left(\mathbf{n}_j\right)\) are:
\begin{equation}
\label{eq:parc_der_perc}
\begin{split}
\frac{\partial f}{\partial n_{1j}} &= \frac{\partial f}{\partial n_{2j}} = 1,\\
\frac{\partial g}{\partial n_{1j}} &= -\frac{c_1\left|\widehat{\phi}_j^{(2)}\right|}{2n_{1j}^2\sqrt{c_1n_{1j}^{-1} + c_2n_{2j}^{-1}}},\\
\frac{\partial g}{\partial n_{2j}} &= -\frac{c_2\left|\widehat{\phi}_j^{(2)}\right|}{2n_{2j}^2\sqrt{c_1n_{1j}^{-1} + c_2n_{2j}^{-1}}}.
\end{split}
\end{equation}

Using these results in \eqref{eq:opt_sys} results in the system
\begin{align}
\beta c_1\left|\widehat{\phi}_j^{(2)}\right| &= 2n_{1j}^2\sqrt{c_1n_{1j}^{-1} + c_2n_{2j}^{-1}},\label{eq:kkt2.4}\\
\beta c_2\left|\widehat{\phi}_j^{(2)}\right| &= 2n_{2j}^2\sqrt{c_1n_{1j}^{-1} + c_2n_{2j}^{-1}}.\label{eq:kkt2.5}\\
\left|\widehat{\phi}_j^{(2)}\right|\sqrt{c_1n_{1j}^{-1} + c_2n_{2j}^{-1}}  &= se^*,\label{eq:kkt2.3}
\end{align}

\noindent with \(\beta > 0\) (again, simple inspection of
\eqref{eq:kkt2.4} shows \(\beta=0\) to be impossible).  
Dividing \eqref{eq:kkt2.4} by \eqref{eq:kkt2.5} yields the optimal
sample size ratio,
\begin{equation}
\label{eq:opt_ratio2}
r_{opt} = \frac{n_{1j}}{n_{2j}} = \sqrt{\frac{c1}{c2}} = \frac{s_{1j}}{s_{2j}}\sqrt{1+\left(\widehat{\phi}_j^{(2)}\right)^2}
\end{equation}


The expressions in \eqref{eq:FiellerSE_nocov} and \eqref{eq:opt_ratio2} can be used directly into Algorithm \ref{alg:reps_abs}, so that the adequate sample sizes for obtaining an estimate $\widehat{\phi}_j^{(2)}$ with a standard error controlled at a given threshold $se^*$ can be iteratively generated.

%
%

\section{ Quick Use Guide for Running Experiments}

\subsection{Introduction}
This document provides a quick use guide to the methodology described in the main text of ``\textit{Sample size estimation for power and accuracy in the experimental comparison of algorithms}''. This guide covers the main concepts needed for using the proposed methodology to design and run comparative experiments on algorithms, with references to the specific sections of the main text where the concepts are introduced, as well as instructions on how to perform each step using the routines implemented in the \texttt{CAISEr} package \cite{caiser}. 

Two use cases are covered, and for each one we prepared a description that is independent of the other two. While this resulted in some amount of repetition, we opted for this strategy so as to allow readers to select the section most similar to their intended use as a standalone ``tutorial'' for the methodology. The use cases described here are not intended as an exhaustive list of possible applications, but instead as an example of two experimental situations in which we believe the proposed methodology can be useful.
\newpage

\subsection{Use Case 1: Full Experimental Setup}
The first use case covers the situation in which the experimenter wishes to design a comparative experiment essentially from scratch: he or she has a certain problem class of interest; two algorithms to be compared; and potential access to a certain number of problem instances, or maybe to an instance generator. The researcher is interested in determining the smallest number of instances required (for achieving certain statistical properties in the comparison) as well as the optimal number of runs of each algorithm on each instance. The need for these calculations arise, for example, in computationally expensive contexts \cite{Jones1998}. Typical cases include the comparison of algorithms for problem classes in which evaluation of the objective or constraint functions are costly, e.g., applied optimization based on numeric models \cite{Mori2015}; or situations in which the algorithms need to run for a very large number of iterations, e.g., when trying to solve very large instances of classic problems in operations research \cite{Maravilha2018}.

\subsection{Number of instances}
To use the proposed methodology for determining a full experimental design, the researcher must be able to provide some input values related to the desired statistical properties. First, to estimate the smallest number of instances required for the experiment, the following information must be provided:
\begin{itemize}
\item Desired significance level, $\alpha$;
\item Minimally relevant effect size (MRES), $d^*$;
\item Desired statistical power, $\pi^*$;
\item Direction of the alternative hypothesis (one sided or two-sided).
\end{itemize}

Some suggestions for setting reasonable values for the first three points are provided in Section 4.6 (``\textit{Defining reasonable experimental parameters}'') of the main text, and a detailed description of these concepts is available in Section 3 (``\textit{Relevant statistical concepts}''). The definition of the direction of the alternative hypotheses is related to the kind of question being asked in the experiment: general questions of the type ``is one of these methods better than the other in terms of mean performance'' should employ a two-sided alternative; whereas questions where there is a specific direction of interest (such as ``is method \textbf{A} better than method \textbf{B} in terms of mean performance'') can benefit from using a one-sided alternative hypothesis, which allows using smaller sample sizes to achieve the same statistical power. 

Once the researcher has defined values for each point above, estimating the smallest required number of instances to obtain the desired properties is just a matter of using the chosen values into expressions (23) or (24) of the main text (Section 4.2, ``\textit{Estimating the number of instances}''). 

If the \texttt{CAISEr} package is used, this can be done as follows. Here we assumed that the researcher has set $\alpha = 0.05$, $\pi^* = 0.85$ and $d^* = 0.5$, for a two-sided alternative hypothesis. In the \texttt{R} environment, the smallest number of instances required for the experiment can be estimated as follows:
\newpage
\begin{Verbatim}[frame=single, rulecolor=\color{black}]
> install.packages("CAISEr") # To install from CRAN
> library(CAISEr)            # Load package

# Calculate required number of instances, considering the t-test
> Ncalc <- calc_instances(power       = 0.85, 
+                         d           = 0.5, 
+                         sig.level   = 0.05, 
+                         alternative = "two.sided", 
+                         test.type   = "t.test")
> Ncalc$ninstances
[1] 38
\end{Verbatim}

\noindent that is, for the experiment to achieve the desired statistical properties (namely, to have an $85\%$ probability of detecting differences at least $0.5$ standard deviations with a significance level $0.05$) it would require at least $38$ instances, considering that the assumptions of the t-test will be valid. If the researcher wants to be more conservative and not assume normality of the sampling distribution of means, he or she could opt for calculating the sample size considering, e.g., the use of a Wilcoxon Signed-ranks test, as discussed in Section 4.4.2 (``Nonparametric tests of hypotheses'') of the main text. As mentioned in the main text, the number of instances in this case can be estimated by multiplying the quantity calculated above by $1.16$ and rounding up the resulting number of instances; or, if the \texttt{CAISEr} package is used:

\begin{Verbatim}[frame=single, rulecolor=\color{black}]
# Calculate required number of instances, considering Wilcoxon's test
> Ncalc <- calc_instances(power       = 0.85, 
+                         d           = 0.5, 
+                         sig.level   = 0.05, 
+                         alternative = "two.sided", 
+                         test.type   = "wilcoxon")
> Ncalc$ninstances
[1] 45
\end{Verbatim}

\noindent which, as explained, requires a larger number of instances to be able to guarantee the same statistical power.

\subsection{Number of repetitions}
After the number of instances is determined, the user can run the experiment using the proposed methodology to iteratively estimate the required number of runs for each algorithm on each instance, as described in Section 4.1 (``\textit{Estimating the number of repetitions}'') of the main text. For that, the following quantities must be defined:

\begin{itemize}
\item Accuracy threshold for the estimation, $se^*$;
\item Initial number of repetitions, $n_0$;
\item Maximum allowed number of repetitions, $n_{max}$;
\item Type of paired difference to use (simple differences or percent gains)
\end{itemize}

Suggestions for setting the first three values are also provided in Section 4.6 (``\textit{Defining reasonable experimental parameters}'') of the main text, and a detailed description of these concepts is available in Section 4.1 (``\textit{Estimating the number of repetitions}''). The determination of which type of paired difference to use depends on the way the researcher wants to quantify the differences in performance between the two algorithms (namely, mean differences or mean percent gains).

After these values are chosen, the researcher is able to, for each instance included in the experiment (ideally equal to or greater than the number calculated earlier), run each algorithm a number of times iteratively determined so as to obtain estimates of performance difference with the predefined precision $se^*$. This can be done by following the sampling procedure illustrated in Algorithm 1 (Section 4.1, ``\textit{Estimating the number of repetitions}'') of the main text. 

Using the implementation available in the \texttt{CAISEr} package to perform this step on a single instance is straightforward. To illustrate it, we borrow an example from the documentation of function \texttt{optim} of the standard \texttt{R} package \texttt{stats}, which is concerned with optimizing a 21-city Traveling Salesman instance using Simulated Annealing. We assume, for the sake of simplicity, that the experiment was concerned with comparing two configurations of Simulated Annealing (varying only the $Temp$ parameter). Both the algorithm runner and the instance are provided as examples in the documentation of the \texttt{CAISEr} package. Supposing the researcher wants an accuracy threshold  $se^* = 0.01$ on the percent gains of one configuration over the other; and that he or she sets $n_0 = 20,~n_{max} = 200$, we have:

\begin{Verbatim}[frame=single, rulecolor=\color{black}]
# Run both algorithm configurations on a single problem instance, 
# with number of repetitions determined according to the 
# proposed methodology
> algorithm1 <- list(FUN  = "my.SANN", alias = "algo1",
+                    Temp = 2000, budget = 10000)

> algorithm2 <- list(FUN  = "my.SANN", alias = "algo2",
+                    Temp = 4000, budget = 10000)

> instance <- list(FUN = "TSP.dist", mydist = datasets::eurodist)

> my.reps  <- calc_nreps2(instance, algorithm1, algorithm2,
+                         se.max = 0.01, dif = "perc",
+                         method = "param", seed = 1234,
+                         nstart = 20)

# Get summary
> cat("n1j   =", my.reps$n1j, "\nn2j   =", my.reps$n2j,
+     "\nphi_j =", my.reps$phi.est, "\nse    =", my.reps$se)

n1j   = 48 
n2j   = 41 
phi_j = 0.05267972 
se    = 0.009958074
\end{Verbatim}

\noindent which shows that the second configuration (sampled 41 times) yielded routes that presented, on average, $5.27\%$ longer distances when compared to the first (sampled 48 times) for that particular instance. As desired, the standard error of estimation was just below $0.01$. The complete set of results can be extracted from the output structure \verb|my.reps|.

To run the complete experiment using the proposed methodology, the researcher simply needs to sample each algorithm on each instance (again, ideally at least as many instances as determined in the first part) using the proposed method for estimating the number of repetitions. This can be achieved in the \texttt{CAISEr} package by repeatedly calling \verb|calc_nreps2()|, as shown above, for each instance; or by using the wrapper function \verb|run_experiment()|, provided in the package.
\newpage

\subsection{Use Case II: Using a Predefined Set of Benchmark Instances}

The second use case covers the situation in which the experimenter wishes to compare two algorithms for a given problem class using a predefined, fixed-size set of benchmark instances. This is a common situation in several problem domains (e.g., scheduling problems \cite{Vallada2011} or multiobjective optimization \cite{Huband2006}). In these cases, the researcher does not need to calculate the number of instances for the experiment, which is predefined for the experiment. The proposed methodology can nonetheless be valuable as an experimental tool. Two elements of the experimental design that can be provided in these cases are:

\begin{itemize}
	\item Determination of the number of runs for each algorithm on each instance.
	\item Determination of the expected power of the experiment, i.e., its ability to detect differences larger than a given minimally relevant effect size (MRES).
\end{itemize}

\subsection{Number of repetitions}
To perform the experiment using the proposed methodology to iteratively estimate the required number of runs for each algorithm on each instance, the researcher must define the following quantities, as presented in Section 4.1 (``\textit{Estimating the number of repetitions}'') of the main text:

\begin{itemize}
	\item Accuracy threshold for the estimation, $se^*$;
	\item Initial number of repetitions, $n_0$;
	\item Maximum allowed number of repetitions, $n_{max}$;
	\item Type of paired difference to use (simple differences or percent gains)
\end{itemize}

Suggestions for setting the first three values are provided in Section 4.6 (``\textit{Defining reasonable experimental parameters}'') of the main text. The determination of which type of paired difference to use depends on the way the researcher wants to quantify the differences in performance between the two algorithms (namely, mean differences or mean percent gains), and a detailed description of these concepts is available in Section 4.1 (``\textit{Estimating the number of repetitions}'').

After these values are determined, the researcher is able to, for each instance of the benchmark set, run each algorithm a number of times iteratively determined so as to obtain estimates of performance difference with the predefined precision $se^*$. This can be done by following the sampling procedure illustrated in the Algorithm 1 of the main text. 

Using the implementation available in the \texttt{R} package \texttt{CAISEr} to adaptively sample two algorithms on a single instance is straightforward. To illustrate it, we borrow an example from the documentation of function \texttt{optim} of the standard \texttt{R} package \texttt{stats}, which is concerned with optimizing a 21-city Traveling Salesman instance using Simulated Annealing. We assume, for the sake of simplicity, that the experiment was concerned with comparing two configurations of Simulated Annealing (varying only the $Temp$ parameter). Both the algorithm runner and the instance are provided as examples in the documentation of the \texttt{CAISEr} package. Supposing the researcher wants an accuracy threshold  $se^* = 0.01$ on the percent gains of one configuration over the other; and that he or she sets $n_0 = 20,~n_{max} = 200$, we have:

\begin{Verbatim}[frame=single, rulecolor=\color{black}]
# Run both algorithm configurations on a single problem instance, 
# with number of repetitions determined according to the 
# proposed methodology
> algorithm1 <- list(FUN  = "my.SANN", alias = "algo1",
+                    Temp = 2000, budget = 10000)

> algorithm2 <- list(FUN  = "my.SANN", alias = "algo2",
+                    Temp = 4000, budget = 10000)

> instance <- list(FUN = "TSP.dist", mydist = datasets::eurodist)

> my.reps  <- calc_nreps2(instance, algorithm1, algorithm2,
+                         se.max = 0.01, dif = "perc",
+                         method = "param", seed = 1234,
+                         nstart = 20)

# Get summary
> cat("n1j   =", my.reps$n1j, "\nn2j   =", my.reps$n2j,
+     "\nphi_j =", my.reps$phi.est, "\nse    =", my.reps$se)

n1j   = 48 
n2j   = 41 
phi_j = 0.05267972 
se    = 0.009958074
\end{Verbatim}

\noindent which shows that the second configuration (sampled 41 times) yielded routes that presented, on average, $5.27\%$ longer distances when compared to the first (sampled 48 times) for that particular instance. As desired, the standard error of estimation was just below $0.01$. The complete set of results can be extracted from the output structure \verb|my.reps|.

To run the complete experiment using the proposed methodology, the researcher simply needs to sample each algorithm on each instance (again, ideally at least as many instances as determined in the first part) using the proposed method for estimating the number of repetitions. This can be achieved in the \texttt{CAISEr} package by repeatedly calling \verb|calc_nreps2()|, as shown above, for each instance; or by using the wrapper function \verb|run_experiment()|, provided in the package.

\subsection{Power estimation}
When a predefined test set needs to be used for the experiment, the researcher has no control over the number of instances to be employed. In these cases, the method used for estimating the number of instances in the proposed methodology can be used instead to estimate the statistical power of the experiment, as explained in Section 4.5 (``\textit{The case of predefined $N$}'') of the main text. To estimate the power of a given experiment \textit{a priori}, the researcher must provide the following information:
\begin{itemize}
	\item Number of instances to be used, $N$;
	\item Desired significance level, $\alpha$;
	\item Minimally relevant effect size (MRES), $d^*$;
	\item Direction of the alternative hypothesis (one sided or two-sided).
\end{itemize}

In the case of a fixed benchmark set, the value of $N$ is simply the number of available instances in the set. Some suggestions for setting reasonable values for the second and third points are provided in Section 4.6 (``\textit{Defining reasonable experimental parameters}'') of the main text, and a detailed description of these concepts is available in Section 3 (``\textit{Relevant statistical concepts}''). The definition of the direction of the alternative hypotheses is related to the kind of question being asked in the experiment: general questions of the type ``is one of these methods better than the other in terms of mean performance'' should employ a two-sided alternative; whereas questions where there is a specific direction of interest (such as ``is method \textbf{A} better than method \textbf{B} in terms of mean performance'') can benefit from using a one-sided alternative hypothesis, which allows using smaller sample sizes to achieve the same statistical power. 

After these values are defined, the researcher can simply compute the integral in Equation (22) (Section 4.2, ``\textit{Estimating the number of instances}'', of the main text) to compute the power of the experiment, i.e., its sensitivity to detect differences of at least $d^*$ under significance level $\alpha$. Using the implementation provided in the \texttt{CAISEr} package, this can be done quite easily. Assuming the researcher has a benchmark set composed of $N = 100$ instances, and that the other experimental parameters were set as $\alpha = 0.01,~d^* = 0.25$, and that he or she intends to use a test with a one-sided alternative, we have:

\begin{Verbatim}[frame=single, rulecolor=\color{black}]
> my.power <- calc_instances(ninstances  = 100, 
+                            d           = 0.25, 
+                            sig.level   = 0.01, 
+                            alternative = "one.sided", 
+                            test.type   = "t.test")
> 
> my.power$power
[1] 0.5554571
\end{Verbatim}

\noindent that is, a 100-instance benchmark set will have a power of $\pi^* \approxeq 0.55$ to detect standardized differences of $d^* = 0.25$ under a significance level of $\alpha = 0.01$ using a one-sided alternative hypothesis (the power is even lower if a two-sided alternative is used). To obtain a more detailed power profile for the experiment the researcher can iterate over different possible values of $d^*$ to obtain a power curve for this experiment, as outlined in Section 4.5 (``\textit{The case of predefined $N$}'') of the main text. Using \texttt{CAISEr} yields a summary table and a power curve:
\newpage
\begin{Verbatim}[frame=single, rulecolor=\color{black}]
> my.pc <- calc_power_curve(ninstances  = 100,
+                           sig.level   = 0.01,
+                           alternative = "one.sided",
+                           test.type   = "t.test",
+                           d.range     = c(0.05, 0.5),
+                           npoints     = 300)

> plot(my.pc, highlights = c(.25, .5, .8, .95))
#====================================
Number of points:  300
power = 0.25 for d = 0.17
power = 0.5  for d = 0.24
power = 0.8  for d = 0.32
power = 0.95 for d = 0.4
#====================================
\end{Verbatim}
\begin{figure}[ht]
	\centering
	\includegraphics[width=0.7\textwidth]{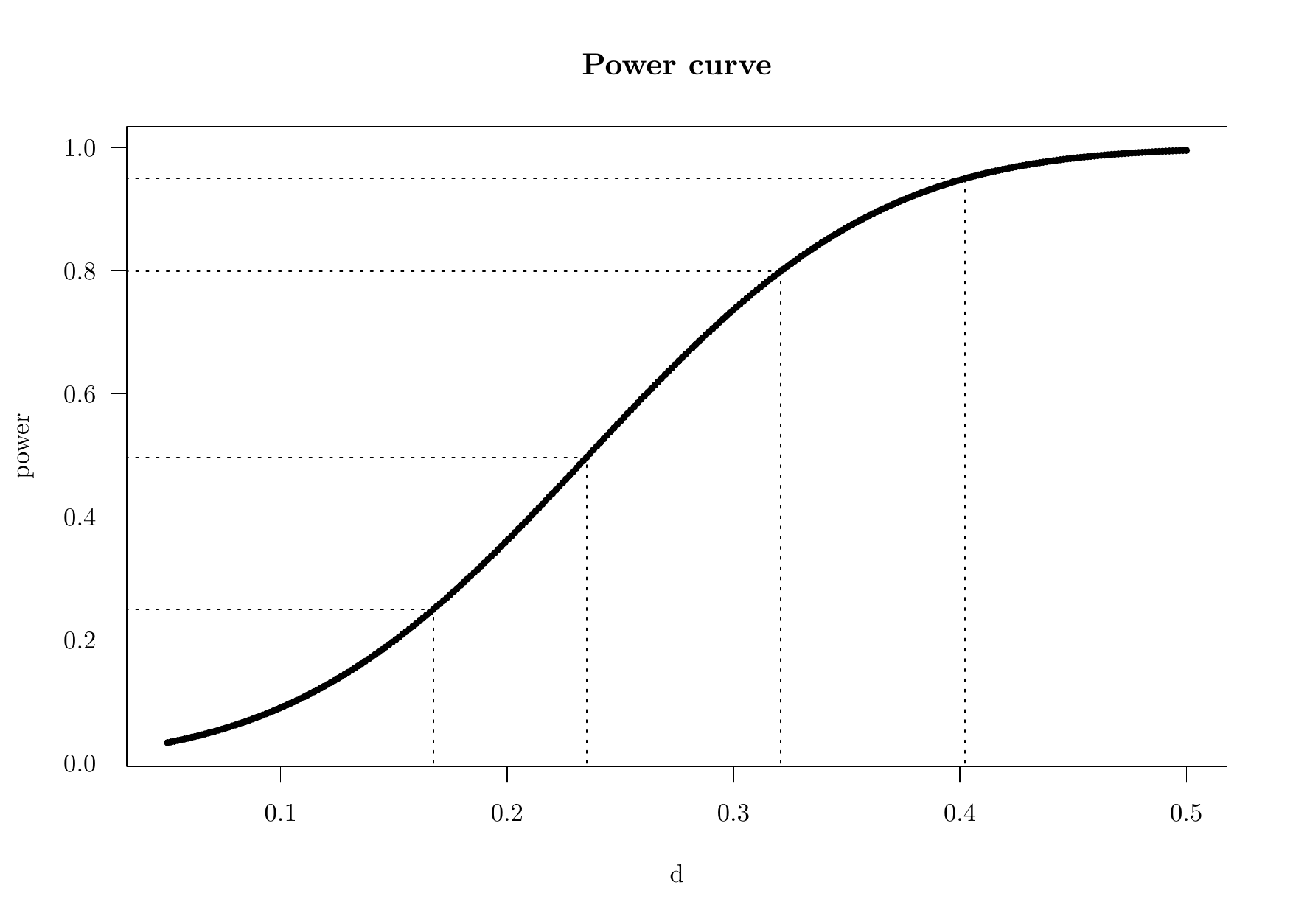}
	\caption{Power curve for the 100-instance example.}
	\label{fig:pwrcurve}
\end{figure}
This result informs the researcher regarding which effect sizes the experiment would actually be capable of identifying. This information can be useful, e.g., in determining how to interpret eventual failures to reject the null hypothesis\footnote{Understanding that failure to detect differences is different from establishing equivalence.}, enabling the researcher to infer, at least qualitatively, upper bounds for possibly-undetected differences in mean performance of the algorithms.

\section{Results Table and plots for Experiment 2}

\renewcommand{\arraystretch}{1.3}
\footnotesize
\begin{longtable}{c|l|c|c|c|c}
	\caption{Summary of results obtained in Experiment 2.}\\
	\hline
	&Instance&$\widehat{\phi}_j$ & $\widehat{se}_{\widehat{\phi}_j}$&$n_{1j}$ & $n_{2j}$\\
	\hline
	\endfirsthead
	\multicolumn{6}{c}%
	{\tablename\ \thetable\ -- \textit{Continued from previous page}} \\
	\hline
	&Instance&$\widehat{\phi}_j$ & $\widehat{se}_{\widehat{\phi}_j}$&$n_{1j}$ & $n_{2j}$\\
	\hline
	\endhead
	\hline \multicolumn{6}{r}{\textit{Continued on next page}} \\
	\endfoot
	\hline
	\endlastfoot
1 & I\_100\_10\_S\_1-124\_1 & -0.23 & 0.05 & 17.00 & 15.00 \\ 
2 & I\_100\_10\_S\_1-124\_2 & -0.18 & 0.04 & 15.00 & 15.00 \\ 
3 & I\_100\_10\_S\_1-49\_1 & -0.25 & 0.04 & 15.00 & 15.00 \\ 
4 & I\_100\_10\_S\_1-49\_2 & -0.32 & 0.05 & 15.00 & 15.00 \\ 
5 & I\_100\_10\_S\_1-9\_1 & -0.28 & 0.04 & 15.00 & 15.00 \\ 
6 & I\_100\_10\_S\_1-9\_2 & -0.32 & 0.04 & 15.00 & 15.00 \\ 
7 & I\_100\_10\_S\_1-99\_1 & -0.24 & 0.05 & 18.00 & 16.00 \\ 
8 & I\_100\_10\_S\_1-99\_2 & -0.25 & 0.04 & 15.00 & 15.00 \\ 
9 & I\_100\_15\_S\_1-124\_1 & -0.30 & 0.05 & 15.00 & 15.00 \\ 
10 & I\_100\_15\_S\_1-124\_2 & -0.27 & 0.04 & 15.00 & 15.00 \\ 
11 & I\_100\_15\_S\_1-49\_1 & -0.31 & 0.05 & 15.00 & 15.00 \\ 
12 & I\_100\_15\_S\_1-49\_2 & -0.27 & 0.04 & 15.00 & 15.00 \\ 
13 & I\_100\_15\_S\_1-9\_1 & -0.39 & 0.05 & 16.00 & 15.00 \\ 
14 & I\_100\_15\_S\_1-9\_2 & -0.39 & 0.03 & 15.00 & 15.00 \\ 
15 & I\_100\_15\_S\_1-99\_1 & -0.25 & 0.04 & 15.00 & 15.00 \\ 
16 & I\_100\_15\_S\_1-99\_2 & -0.35 & 0.04 & 15.00 & 15.00 \\ 
17 & I\_100\_20\_S\_1-124\_1 & -0.34 & 0.04 & 15.00 & 15.00 \\ 
18 & I\_100\_20\_S\_1-124\_2 & -0.35 & 0.04 & 15.00 & 15.00 \\ 
19 & I\_100\_20\_S\_1-49\_1 & -0.39 & 0.04 & 15.00 & 15.00 \\ 
20 & I\_100\_20\_S\_1-49\_2 & -0.38 & 0.05 & 18.00 & 15.00 \\ 
21 & I\_100\_20\_S\_1-9\_1 & -0.48 & 0.05 & 33.00 & 15.00 \\ 
22 & I\_100\_20\_S\_1-9\_2 & -0.50 & 0.05 & 16.00 & 15.00 \\ 
23 & I\_100\_20\_S\_1-99\_1 & -0.36 & 0.05 & 19.00 & 15.00 \\ 
24 & I\_100\_20\_S\_1-99\_2 & -0.35 & 0.05 & 20.00 & 15.00 \\ 
25 & I\_100\_25\_S\_1-124\_1 & -0.40 & 0.05 & 21.00 & 15.00 \\ 
26 & I\_100\_25\_S\_1-124\_2 & -0.40 & 0.04 & 15.00 & 15.00 \\ 
27 & I\_100\_25\_S\_1-49\_1 & -0.49 & 0.05 & 15.00 & 15.00 \\ 
28 & I\_100\_25\_S\_1-49\_2 & -0.47 & 0.05 & 20.00 & 15.00 \\ 
29 & I\_100\_25\_S\_1-9\_1 & -0.57 & 0.05 & 42.00 & 15.00 \\ 
30 & I\_100\_25\_S\_1-9\_2 & -0.57 & 0.05 & 33.00 & 15.00 \\ 
31 & I\_100\_25\_S\_1-99\_1 & -0.44 & 0.05 & 21.00 & 15.00 \\ 
32 & I\_100\_25\_S\_1-99\_2 & -0.43 & 0.04 & 15.00 & 15.00 \\ 
33 & I\_100\_30\_S\_1-124\_1 & -0.43 & 0.04 & 15.00 & 15.00 \\ 
34 & I\_100\_30\_S\_1-124\_2 & -0.49 & 0.05 & 16.00 & 15.00 \\ 
35 & I\_100\_30\_S\_1-49\_1 & -0.48 & 0.05 & 22.00 & 15.00 \\ 
36 & I\_100\_30\_S\_1-49\_2 & -0.55 & 0.05 & 34.00 & 15.00 \\ 
37 & I\_100\_30\_S\_1-9\_1 & -0.57 & 0.05 & 26.00 & 15.00 \\ 
38 & I\_100\_30\_S\_1-9\_2 & -0.58 & 0.05 & 26.00 & 15.00 \\ 
39 & I\_100\_30\_S\_1-99\_1 & -0.45 & 0.04 & 15.00 & 15.00 \\ 
40 & I\_100\_30\_S\_1-99\_2 & -0.45 & 0.05 & 30.00 & 15.00 \\ 
41 & I\_150\_10\_S\_1-124\_1 & -0.13 & 0.04 & 15.00 & 15.00 \\ 
42 & I\_150\_10\_S\_1-124\_2 & -0.15 & 0.03 & 15.00 & 15.00 \\ 
43 & I\_150\_10\_S\_1-49\_1 & -0.19 & 0.04 & 15.00 & 15.00 \\ 
44 & I\_150\_10\_S\_1-49\_2 & -0.21 & 0.04 & 15.00 & 15.00 \\ 
45 & I\_150\_10\_S\_1-9\_1 & -0.24 & 0.03 & 15.00 & 15.00 \\ 
46 & I\_150\_10\_S\_1-9\_2 & -0.31 & 0.03 & 15.00 & 15.00 \\ 
47 & I\_150\_10\_S\_1-99\_1 & -0.13 & 0.04 & 15.00 & 15.00 \\ 
48 & I\_150\_10\_S\_1-99\_2 & -0.21 & 0.03 & 15.00 & 15.00 \\ 
49 & I\_150\_15\_S\_1-124\_1 & -0.25 & 0.05 & 15.00 & 15.00 \\ 
50 & I\_150\_15\_S\_1-124\_2 & -0.21 & 0.05 & 15.00 & 15.00 \\ 
51 & I\_150\_15\_S\_1-49\_1 & -0.31 & 0.04 & 15.00 & 15.00 \\ 
52 & I\_150\_15\_S\_1-49\_2 & -0.25 & 0.04 & 15.00 & 15.00 \\ 
53 & I\_150\_15\_S\_1-9\_1 & -0.41 & 0.05 & 16.00 & 15.00 \\ 
54 & I\_150\_15\_S\_1-9\_2 & -0.39 & 0.05 & 15.00 & 15.00 \\ 
55 & I\_150\_15\_S\_1-99\_1 & -0.26 & 0.03 & 15.00 & 15.00 \\ 
56 & I\_150\_15\_S\_1-99\_2 & -0.25 & 0.04 & 15.00 & 15.00 \\ 
57 & I\_150\_20\_S\_1-124\_1 & -0.30 & 0.04 & 15.00 & 15.00 \\ 
58 & I\_150\_20\_S\_1-124\_2 & -0.30 & 0.05 & 15.00 & 15.00 \\ 
59 & I\_150\_20\_S\_1-49\_1 & -0.41 & 0.05 & 18.00 & 15.00 \\ 
60 & I\_150\_20\_S\_1-49\_2 & -0.36 & 0.03 & 15.00 & 15.00 \\ 
61 & I\_150\_20\_S\_1-9\_1 & -0.46 & 0.04 & 15.00 & 15.00 \\ 
62 & I\_150\_20\_S\_1-9\_2 & -0.52 & 0.05 & 18.00 & 15.00 \\ 
63 & I\_150\_20\_S\_1-99\_1 & -0.31 & 0.04 & 15.00 & 15.00 \\ 
64 & I\_150\_20\_S\_1-99\_2 & -0.24 & 0.04 & 15.00 & 15.00 \\ 
65 & I\_150\_25\_S\_1-124\_1 & -0.41 & 0.04 & 15.00 & 15.00 \\ 
66 & I\_150\_25\_S\_1-124\_2 & -0.37 & 0.05 & 15.00 & 15.00 \\ 
67 & I\_150\_25\_S\_1-49\_1 & -0.38 & 0.03 & 15.00 & 15.00 \\ 
68 & I\_150\_25\_S\_1-49\_2 & -0.40 & 0.04 & 15.00 & 15.00 \\ 
69 & I\_150\_25\_S\_1-9\_1 & -0.53 & 0.05 & 19.00 & 15.00 \\ 
70 & I\_150\_25\_S\_1-9\_2 & -0.49 & 0.05 & 17.00 & 15.00 \\ 
71 & I\_150\_25\_S\_1-99\_1 & -0.31 & 0.04 & 15.00 & 15.00 \\ 
72 & I\_150\_25\_S\_1-99\_2 & -0.32 & 0.05 & 15.00 & 15.00 \\ 
73 & I\_150\_30\_S\_1-124\_1 & -0.39 & 0.05 & 16.00 & 15.00 \\ 
74 & I\_150\_30\_S\_1-124\_2 & -0.42 & 0.05 & 21.00 & 15.00 \\ 
75 & I\_150\_30\_S\_1-49\_1 & -0.43 & 0.04 & 15.00 & 15.00 \\ 
76 & I\_150\_30\_S\_1-49\_2 & -0.43 & 0.03 & 15.00 & 15.00 \\ 
77 & I\_150\_30\_S\_1-9\_1 & -0.52 & 0.05 & 16.00 & 15.00 \\ 
78 & I\_150\_30\_S\_1-9\_2 & -0.59 & 0.05 & 19.00 & 15.00 \\ 
79 & I\_150\_30\_S\_1-99\_1 & -0.38 & 0.05 & 17.00 & 15.00 \\ 
80 & I\_150\_30\_S\_1-99\_2 & -0.36 & 0.05 & 16.00 & 15.00 \\ 
81 & I\_200\_10\_S\_1-124\_1 & -0.23 & 0.05 & 21.00 & 15.00 \\ 
82 & I\_200\_10\_S\_1-124\_2 & -0.19 & 0.04 & 15.00 & 15.00 \\ 
83 & I\_200\_10\_S\_1-49\_1 & -0.18 & 0.02 & 15.00 & 15.00 \\ 
84 & I\_200\_10\_S\_1-49\_2 & -0.14 & 0.04 & 15.00 & 15.00 \\ 
85 & I\_200\_10\_S\_1-9\_1 & -0.19 & 0.03 & 15.00 & 15.00 \\ 
86 & I\_200\_10\_S\_1-9\_2 & -0.17 & 0.03 & 15.00 & 15.00 \\ 
87 & I\_200\_10\_S\_1-99\_1 & -0.14 & 0.03 & 15.00 & 15.00 \\ 
88 & I\_200\_10\_S\_1-99\_2 & -0.17 & 0.03 & 15.00 & 15.00 \\ 
89 & I\_200\_15\_S\_1-124\_1 & -0.26 & 0.03 & 15.00 & 15.00 \\ 
90 & I\_200\_15\_S\_1-124\_2 & -0.21 & 0.03 & 15.00 & 15.00 \\ 
91 & I\_200\_15\_S\_1-49\_1 & -0.31 & 0.04 & 15.00 & 15.00 \\ 
92 & I\_200\_15\_S\_1-49\_2 & -0.25 & 0.04 & 15.00 & 15.00 \\ 
93 & I\_200\_15\_S\_1-9\_1 & -0.35 & 0.05 & 20.00 & 15.00 \\ 
94 & I\_200\_15\_S\_1-9\_2 & -0.28 & 0.03 & 15.00 & 15.00 \\ 
95 & I\_200\_15\_S\_1-99\_1 & -0.30 & 0.04 & 15.00 & 15.00 \\ 
96 & I\_200\_15\_S\_1-99\_2 & -0.21 & 0.04 & 15.00 & 15.00 \\ 
97 & I\_200\_20\_S\_1-124\_1 & -0.25 & 0.05 & 15.00 & 15.00 \\ 
98 & I\_200\_20\_S\_1-124\_2 & -0.26 & 0.03 & 15.00 & 15.00 \\ 
99 & I\_200\_20\_S\_1-49\_1 & -0.37 & 0.03 & 15.00 & 15.00 \\ 
100 & I\_200\_20\_S\_1-49\_2 & -0.33 & 0.04 & 15.00 & 15.00 \\ 
101 & I\_200\_20\_S\_1-9\_1 & -0.37 & 0.04 & 15.00 & 15.00 \\ 
102 & I\_200\_20\_S\_1-9\_2 & -0.42 & 0.04 & 15.00 & 15.00 \\ 
103 & I\_200\_20\_S\_1-99\_1 & -0.27 & 0.04 & 15.00 & 15.00 \\ 
104 & I\_200\_20\_S\_1-99\_2 & -0.27 & 0.03 & 15.00 & 15.00 \\ 
105 & I\_200\_25\_S\_1-124\_1 & -0.35 & 0.05 & 15.00 & 15.00 \\ 
106 & I\_200\_25\_S\_1-124\_2 & -0.26 & 0.03 & 15.00 & 15.00 \\ 
107 & I\_200\_25\_S\_1-49\_1 & -0.39 & 0.03 & 15.00 & 15.00 \\ 
108 & I\_200\_25\_S\_1-49\_2 & -0.30 & 0.03 & 15.00 & 15.00 \\ 
109 & I\_200\_25\_S\_1-9\_1 & -0.45 & 0.04 & 15.00 & 15.00 \\ 
110 & I\_200\_25\_S\_1-9\_2 & -0.37 & 0.04 & 15.00 & 15.00 \\ 
111 & I\_200\_25\_S\_1-99\_1 & -0.30 & 0.03 & 15.00 & 15.00 \\ 
112 & I\_200\_25\_S\_1-99\_2 & -0.33 & 0.04 & 15.00 & 15.00 \\ 
113 & I\_200\_30\_S\_1-124\_1 & -0.30 & 0.03 & 15.00 & 15.00 \\ 
114 & I\_200\_30\_S\_1-124\_2 & -0.33 & 0.04 & 15.00 & 15.00 \\ 
115 & I\_200\_30\_S\_1-49\_1 & -0.42 & 0.04 & 15.00 & 15.00 \\ 
116 & I\_200\_30\_S\_1-49\_2 & -0.37 & 0.05 & 15.00 & 15.00 \\ 
117 & I\_200\_30\_S\_1-9\_1 & -0.52 & 0.03 & 15.00 & 15.00 \\ 
118 & I\_200\_30\_S\_1-9\_2 & -0.51 & 0.04 & 15.00 & 15.00 \\ 
119 & I\_200\_30\_S\_1-99\_1 & -0.35 & 0.03 & 15.00 & 15.00 \\ 
120 & I\_200\_30\_S\_1-99\_2 & -0.36 & 0.03 & 15.00 & 15.00 \\ 
121 & I\_250\_10\_S\_1-124\_1 & -0.15 & 0.02 & 15.00 & 15.00 \\ 
122 & I\_250\_10\_S\_1-124\_2 & -0.14 & 0.02 & 15.00 & 15.00 \\ 
123 & I\_250\_10\_S\_1-49\_1 & -0.15 & 0.03 & 15.00 & 15.00 \\ 
124 & I\_250\_10\_S\_1-49\_2 & -0.17 & 0.02 & 15.00 & 15.00 \\ 
125 & I\_250\_10\_S\_1-9\_1 & -0.16 & 0.02 & 15.00 & 15.00 \\ 
126 & I\_250\_10\_S\_1-9\_2 & -0.23 & 0.04 & 15.00 & 15.00 \\ 
127 & I\_250\_10\_S\_1-99\_1 & -0.17 & 0.02 & 15.00 & 15.00 \\ 
128 & I\_250\_10\_S\_1-99\_2 & -0.18 & 0.02 & 15.00 & 15.00 \\ 
129 & I\_250\_15\_S\_1-124\_1 & -0.21 & 0.03 & 15.00 & 15.00 \\ 
130 & I\_250\_15\_S\_1-124\_2 & -0.19 & 0.03 & 15.00 & 15.00 \\ 
131 & I\_250\_15\_S\_1-49\_1 & -0.22 & 0.02 & 15.00 & 15.00 \\ 
132 & I\_250\_15\_S\_1-49\_2 & -0.26 & 0.04 & 15.00 & 15.00 \\ 
133 & I\_250\_15\_S\_1-9\_1 & -0.27 & 0.03 & 15.00 & 15.00 \\ 
134 & I\_250\_15\_S\_1-9\_2 & -0.25 & 0.04 & 15.00 & 15.00 \\ 
135 & I\_250\_15\_S\_1-99\_1 & -0.24 & 0.04 & 15.00 & 15.00 \\ 
136 & I\_250\_15\_S\_1-99\_2 & -0.17 & 0.03 & 15.00 & 15.00 \\ 
137 & I\_250\_20\_S\_1-124\_1 & -0.19 & 0.04 & 15.00 & 15.00 \\ 
138 & I\_250\_20\_S\_1-124\_2 & -0.28 & 0.03 & 15.00 & 15.00 \\ 
139 & I\_250\_20\_S\_1-49\_1 & -0.24 & 0.02 & 15.00 & 15.00 \\ 
140 & I\_250\_20\_S\_1-49\_2 & -0.27 & 0.03 & 15.00 & 15.00 \\ 
141 & I\_250\_20\_S\_1-9\_1 & -0.32 & 0.04 & 15.00 & 15.00 \\ 
142 & I\_250\_20\_S\_1-9\_2 & -0.34 & 0.04 & 15.00 & 15.00 \\ 
143 & I\_250\_20\_S\_1-99\_1 & -0.26 & 0.03 & 15.00 & 15.00 \\ 
144 & I\_250\_20\_S\_1-99\_2 & -0.24 & 0.03 & 15.00 & 15.00 \\ 
145 & I\_250\_25\_S\_1-124\_1 & -0.24 & 0.05 & 16.00 & 15.00 \\ 
146 & I\_250\_25\_S\_1-124\_2 & -0.28 & 0.04 & 15.00 & 15.00 \\ 
147 & I\_250\_25\_S\_1-49\_1 & -0.31 & 0.03 & 15.00 & 15.00 \\ 
148 & I\_250\_25\_S\_1-49\_2 & -0.30 & 0.03 & 15.00 & 15.00 \\ 
149 & I\_250\_25\_S\_1-9\_1 & -0.37 & 0.03 & 15.00 & 15.00 \\ 
150 & I\_250\_25\_S\_1-9\_2 & -0.39 & 0.03 & 15.00 & 15.00 \\ 
151 & I\_250\_25\_S\_1-99\_1 & -0.24 & 0.03 & 15.00 & 15.00 \\ 
152 & I\_250\_25\_S\_1-99\_2 & -0.29 & 0.03 & 15.00 & 15.00 \\ 
153 & I\_250\_30\_S\_1-124\_1 & -0.27 & 0.03 & 15.00 & 15.00 \\ 
154 & I\_250\_30\_S\_1-124\_2 & -0.33 & 0.04 & 15.00 & 15.00 \\ 
155 & I\_250\_30\_S\_1-49\_1 & -0.30 & 0.03 & 15.00 & 15.00 \\ 
156 & I\_250\_30\_S\_1-49\_2 & -0.35 & 0.04 & 15.00 & 15.00 \\ 
157 & I\_250\_30\_S\_1-9\_1 & -0.43 & 0.05 & 15.00 & 15.00 \\ 
158 & I\_250\_30\_S\_1-9\_2 & -0.40 & 0.03 & 15.00 & 15.00 \\ 
159 & I\_250\_30\_S\_1-99\_1 & -0.31 & 0.03 & 15.00 & 15.00 \\ 
160 & I\_250\_30\_S\_1-99\_2 & -0.33 & 0.02 & 15.00 & 15.00 \\ 
161 & I\_50\_10\_S\_1-124\_1 & -0.30 & 0.05 & 16.00 & 16.00 \\ 
162 & I\_50\_10\_S\_1-124\_2 & -0.27 & 0.05 & 15.00 & 15.00 \\ 
163 & I\_50\_10\_S\_1-49\_1 & -0.31 & 0.05 & 16.00 & 15.00 \\ 
164 & I\_50\_10\_S\_1-49\_2 & -0.37 & 0.05 & 40.00 & 15.00 \\ 
165 & I\_50\_10\_S\_1-9\_1 & -0.37 & 0.05 & 35.00 & 15.00 \\ 
166 & I\_50\_10\_S\_1-9\_2 & -0.42 & 0.04 & 15.00 & 15.00 \\ 
167 & I\_50\_10\_S\_1-99\_1 & -0.29 & 0.05 & 30.00 & 18.00 \\ 
168 & I\_50\_10\_S\_1-99\_2 & -0.32 & 0.04 & 15.00 & 15.00 \\ 
169 & I\_50\_15\_S\_1-124\_1 & -0.31 & 0.05 & 29.00 & 21.00 \\ 
170 & I\_50\_15\_S\_1-124\_2 & -0.42 & 0.05 & 31.00 & 15.00 \\ 
171 & I\_50\_15\_S\_1-49\_1 & -0.44 & 0.05 & 23.00 & 15.00 \\ 
172 & I\_50\_15\_S\_1-49\_2 & -0.43 & 0.05 & 45.00 & 15.00 \\ 
173 & I\_50\_15\_S\_1-9\_1 & -0.63 & 0.05 & 75.00 & 15.00 \\ 
174 & I\_50\_15\_S\_1-9\_2 & -0.54 & 0.05 & 21.00 & 15.00 \\ 
175 & I\_50\_15\_S\_1-99\_1 & -0.43 & 0.05 & 33.00 & 15.00 \\ 
176 & I\_50\_15\_S\_1-99\_2 & -0.44 & 0.05 & 29.00 & 15.00 \\ 
177 & I\_50\_20\_S\_1-124\_1 & -0.49 & 0.05 & 19.00 & 15.00 \\ 
178 & I\_50\_20\_S\_1-124\_2 & -0.49 & 0.05 & 21.00 & 15.00 \\ 
179 & I\_50\_20\_S\_1-49\_1 & -0.51 & 0.05 & 20.00 & 15.00 \\ 
180 & I\_50\_20\_S\_1-49\_2 & -0.50 & 0.04 & 15.00 & 15.00 \\ 
181 & I\_50\_20\_S\_1-9\_1 & -0.65 & 0.05 & 68.00 & 15.00 \\ 
182 & I\_50\_20\_S\_1-9\_2 & -0.63 & 0.05 & 25.00 & 15.00 \\ 
183 & I\_50\_20\_S\_1-99\_1 & -0.50 & 0.05 & 24.00 & 15.00 \\ 
184 & I\_50\_20\_S\_1-99\_2 & -0.48 & 0.05 & 26.00 & 15.00 \\ 
185 & I\_50\_25\_S\_1-124\_1 & -0.64 & 0.05 & 21.00 & 15.00 \\ 
186 & I\_50\_25\_S\_1-124\_2 & -0.58 & 0.05 & 39.00 & 15.00 \\ 
187 & I\_50\_25\_S\_1-49\_1 & -0.56 & 0.05 & 27.00 & 15.00 \\ 
188 & I\_50\_25\_S\_1-49\_2 & -0.56 & 0.05 & 35.00 & 15.00 \\ 
189 & I\_50\_25\_S\_1-9\_1 & -0.66 & 0.05 & 26.00 & 15.00 \\ 
190 & I\_50\_25\_S\_1-9\_2 & -0.66 & 0.05 & 33.00 & 15.00 \\ 
191 & I\_50\_25\_S\_1-99\_1 & -0.55 & 0.05 & 40.00 & 16.00 \\ 
192 & I\_50\_25\_S\_1-99\_2 & -0.53 & 0.05 & 18.00 & 15.00 \\ 
193 & I\_50\_30\_S\_1-124\_1 & -0.69 & 0.05 & 21.00 & 15.00 \\ 
194 & I\_50\_30\_S\_1-124\_2 & -0.70 & 0.05 & 48.00 & 15.00 \\ 
195 & I\_50\_30\_S\_1-49\_1 & -0.61 & 0.05 & 15.00 & 15.00 \\ 
196 & I\_50\_30\_S\_1-49\_2 & -0.67 & 0.05 & 15.00 & 15.00 \\ 
197 & I\_50\_30\_S\_1-9\_1 & -0.67 & 0.05 & 33.00 & 15.00 \\ 
198 & I\_50\_30\_S\_1-9\_2 & -0.66 & 0.05 & 38.00 & 15.00 \\ 
199 & I\_50\_30\_S\_1-99\_1 & -0.64 & 0.05 & 22.00 & 15.00 \\ 
200 & I\_50\_30\_S\_1-99\_2 & -0.68 & 0.05 & 19.00 & 15.00 \\
			\hline
\end{longtable}

\begin{figure}[ht]
	\centering
	\includegraphics[width = .7\textwidth]{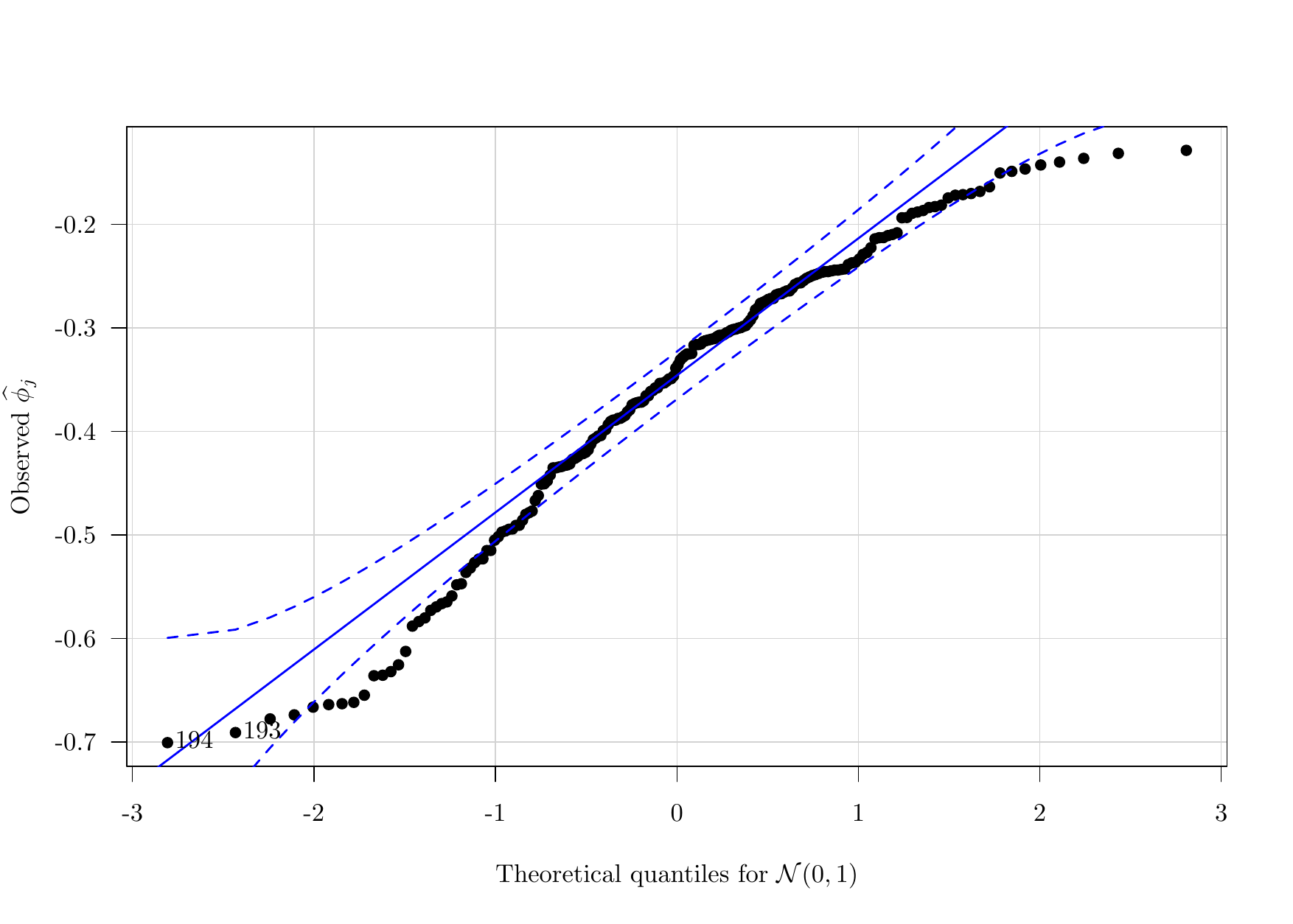}
	\caption{Normal quantile-quantile plot for observations $\widehat{\phi}_j$ in Experiment 2. While there is some indication of small deviations at the tails, the sample size ($N=200$) is large enough to guarantee that the sampling distribution of the means will be close enough to the normal for the t-test to maintain its statistical properties.}
	\label{fig:qqplot2}
\end{figure}

\begin{figure}[ht]
	\centering
	\includegraphics[width = \textwidth]{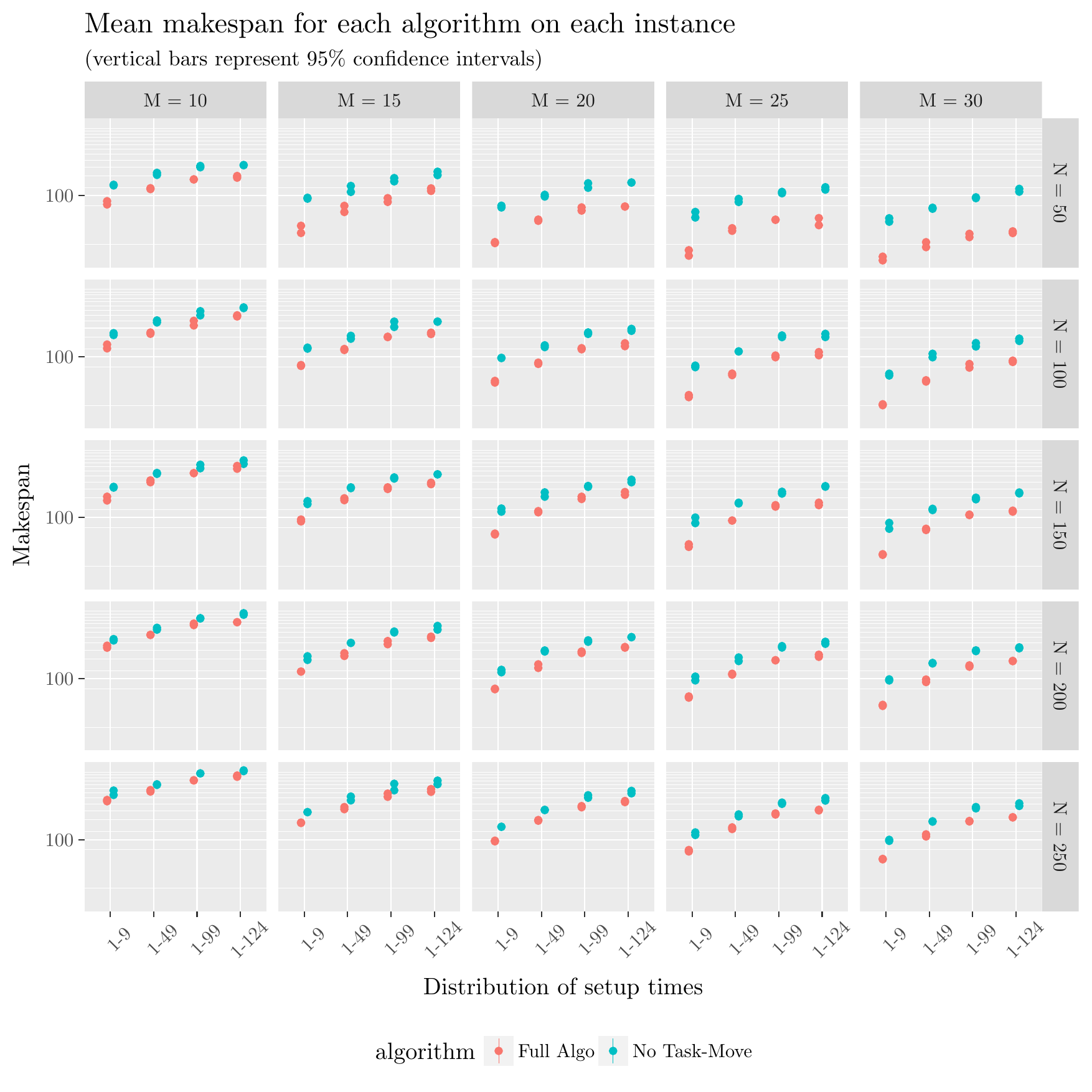}
	\caption{Mean makespan for each algorithm on each instance. Vertical bars represent $95\%$ confidence intervals (invisible if smaller than point size). Notice that there are two instances for each tuple <M, N, SetupTimes> (overlapping in most plots)}
	\label{fig:TM_effect3}
\end{figure}

\bibliographystyle{spmpsci}      